\documentclass[a4paper,fleqn]{cas-sc}

\usepackage[numbers]{natbib}

\usepackage{url}
\usepackage{bm}
\newcommand\Rey{\mbox{\textit{Re}}}  
\newcommand\Ray{\mbox{\textit{Ra}}}  
\newcommand\Prn{\mbox{\textit{Pr}}}  
\newcommand\Nus{\mbox{\textit{Nu}}}  
\usepackage{caption}
\usepackage{subcaption}
\usepackage{soul}

\newcommand{\norm}[1]{\left\lVert#1\right\rVert}

\def\tsc#1{\csdef{#1}{\textsc{\lowercase{#1}}\xspace}}
\tsc{WGM}

\begin{document}
\let\WriteBookmarks\relax

\shorttitle{Coarse-Mesh GNNs for Predicting Sparse Observables in Parameterized Unsteady Flows}
\shortauthors{B. Ramos et~al.}

%
\title [mode = title]{
Predicting Sparse Observables in Parameterized Unsteady Flows Using Coarse-Mesh Graph Neural Networks
}



\author[1]{Brener Ramos}[type=editor,
                        bioid=1,
                        orcid=0000-0002-6325-5647
                        ]
\ead{b138091@dac.unicamp.br}
\cortext[cor1]{Corresponding author}
\cormark[1]

\credit{Conceptualization, Data curation, Formal analysis, Investigation, Methodology, Software, Validation, Visualization, Writing – original draft,  Writing – review and editing}

\affiliation[1]{organization={School of Mechanical Engineering, University of Campinas},
                city={Campinas},
                state={São Paulo},
                country={Brazil}}

\author[1]{William {R. Wolf}}[orcid=0000-0001-8207-8466]
\credit{Conceptualization, Funding acquisition, Methodology, Project administration, Resources, Supervision, Writing – original draft, Writing – review and editing}

\author[1]{Tarcísio {C. D. Oliveira}}[orcid=0000-0002-8298-9478]
\credit{Conceptualization, Methodology, Writing – original draft, Writing – review and editing}

\affiliation[2]{organization={ Department of Mechanical, Materials, and Aerospace Engineering, Illinois Institute of Technology},
                city={Chicago},
                state={Illinois},
                country={USA}}
\author[2]{Scott {T. M. Dawson}}[orcid=0000-0002-0020-2097]

\credit{Conceptualization, Methodology, Writing – original draft, Writing – review and editing}




\begin{abstract}
Reconstructing full-order state vectors in high-dimensional nonlinear dynamical systems such as fluid flows has motivated the development of increasingly complex surrogate models. However, engineering applications such as flow control and shape optimization often require only a subset of a system's degrees of freedom or quantities of interest. Consequently, surrogates that recover the entire flow field may not be the most efficient choice for such applications. Motivated by this, we propose a methodology for constructing reduced-order models (ROMs) that bypass full-state reconstruction by deploying graph neural networks (GNNs) directly on coarse meshes, whereas GNN-based flow surrogates are commonly deployed on meshes comparable to those used by numerical solvers. We apply this approach to parameterized 2D and 3D confined cylinder flows as well as an unevenly heated cavity over ranges of Reynolds and Rayleigh numbers, respectively. By training the ROMs with samples across the parameter spaces, the framework effectively generalizes to predict dynamical responses under unseen flow conditions. Model performance is assessed through prediction accuracy at sparse probe locations and for integrated quantities of interest. Our results show that surrogates trained on coarse meshes achieve comparable or superior accuracy to those trained on finer meshes while requiring less training time and memory, even when considering 3D chaotic dynamics. We also propose a novel dynamics-informed adaptive parameter-space sampling strategy to build training datasets guided by frequency spectrum variations, which reduces maximum errors by an order of magnitude relative to implementations utilizing a uniform sampling strategy while having a negligible computational overhead. Leveraging the reduced cost of coarse-mesh training, we further show that multi-step unrolling with noise augmentation becomes computationally practical and significantly improves long-horizon autoregressive predictions. It is shown that the coarse-mesh GNN outperforms various linear and nonlinear baselines such as parametric dynamic mode decomposition, sparse multilayer perceptrons, U-Nets and Fourier neural operators, while demanding a lower computational cost. We show that the reduced dimensionality and accuracy of the resulting models can be exploited for system identification and state estimation from sparse observations, while comparable fine-mesh models are either less accurate or substantially more computationally demanding. Together, these results demonstrate that coarse-mesh representations, adaptive parameter sampling, and noise-augmented unrolled training provide an efficient framework for reduced-order modeling of parameterized unsteady flows without requiring reconstruction at the spatial resolution of the underlying numerical solver.

\end{abstract}



\begin{keywords}
graph neural networks \sep reduced-order modeling \sep dynamical systems \sep computational fluid dynamics
\end{keywords}

\maketitle


\section{Introduction}

Fluid flows are described by nonlinear partial differential equations (PDEs) that rarely admit analytical solutions. Consequently, computational methods have become an important tool for studying fluid dynamics in science and engineering. Applications span aerodynamics \cite{chapman1979, ramos2019, dorgan2023}, bio-fluid dynamics~\cite{liu2005, lee2011, kenjerevs2016}, heat transfer~\cite{cebeci2013, ozicsik2017}, among others, driving sustained investment in advanced computational tools. However, time-resolved simulations of unsteady flows can be prohibitively expensive. This motivates the use of reduced-order models (ROMs), which try to approximate complex physical systems with simpler, computationally efficient surrogates that use a lower dimensional representation, with applications ranging from control \cite{proctor2016, noack2011, rowley2017} to data assimilation \cite{yaqing2007, arcucci2019}.

Extensive research has been conducted in the development of ROMs for fluid flows. Proper orthogonal decomposition (POD) \cite{pod, lumley2012} extracts a set of orthogonal basis functions directly from data, which optimally capture dominant flow features in an energy sense. Galerkin projection \cite{galerkin, zucatti2021, ZucattiClosure} then reduces the governing PDEs onto a low-dimensional system of ordinary differential equations (ODEs). Dynamic mode decomposition (DMD) \cite{schmid2010} extracts spatiotemporal modes from data that capture the dominant frequencies and growth rates of the flow, while the Koopman operator \cite{koopman1931hamiltonian, rowley2009spectral} provides a theoretical foundation for linearizing nonlinear systems through observable transformations.
One should mention that standard POD-Galerkin and DMD formulations inherently struggle to generalize or interpolate along varying system parameters \cite{benner2015survey}. Mitigating this limitation typically requires complex solutions, such as interpolation on Grassmann manifolds \cite{amsallem2008interpolation} or parametric DMD variants \cite{HUHN2023111852}, which add computational overhead and algorithmic complexity.

Resolvent analysis \cite{mckeon2010critical, gomez2016reduced, pickering2021resolvent} can yield reduced-order representations 
by decomposing the linearized Navier-Stokes operator via singular value decomposition into input and output modes. The nonlinear terms are treated as a forcing (input), and the low-rank structure of the resolvent operator provides a compact basis that captures the flow response (output) through the dominant linear amplification mechanisms. These methods, however, are constrained mainly by their linear projections.
Sparse identification of nonlinear dynamics (SINDy) \cite{brunton2016discovering}
identifies governing ordinary differential equations (ODEs) by selecting a sparse combination of candidate nonlinear functions using $L_1$-regularized optimization. Nevertheless, these approaches remain challenged by strongly nonlinear or chaotic flows, spurring interest in neural-network-based surrogates \cite{lui2019construction}, whose nonlinear modeling capabilities and expressivity offer a promising alternative.


Exploration of surrogate models using neural networks in fluid mechanics dates back to the 1990s, where \cite{faller1997unsteady} used a recurrent network to predict unsteady aerodynamics of an airfoil. However, more recently there was a surge in this field due to  higher availability of data, more computing power and development of new algorithms \cite{brunton2020machine}. Physics informed neural networks (PINNs) \cite{raissi2019physics, cai2021physics, rezaei2022mixed, WASSING2024106164} are trained by minimizing a loss that contains the governing equations of a system and possibly only a sparse set of measurements. However, they are also known for their training instability and difficulty in capturing extremely non-linear phenomena \cite{krishnapriyan2021, wang2021}. Neural operators have emerged as a class of deep learning architectures designed to learn mappings between infinite-dimensional function spaces \cite{li2020neural, Lu_2021, li2021fourier}. Although they are theoretically discretization invariant, they often require substantial spatial resolution during training, particularly for problems defined on non-periodic domains. Note that most deep learning frameworks are not mutually exclusive and can be combined, e.g. a physics informed neural operator \cite{li2024physics} or a neural-operator-based correction term for Galerkin ROMs \cite{ivagnes2026data}.
Autoregressive techniques have also been developed to predict the evolution of PDEs from initial conditions \cite{pfaff2021, liu2024multi}, especially in a reduced-order modeling context \cite{lui2019construction, san2019artificial, maulik2021reduced}. Traditionally, these models work by projecting the full-order system into a latent space with nonlinear mappings.
Following this, the temporal dynamics of the latent space variables are modeled, which can be mapped back to give an estimate of the time-evolution of the original full-order system. However, such approaches are often deployed only in a single set of physical parameters, limiting their generalizability to unseen flow regimes \cite{eivazi2020deep, pant2021deep}.


In engineering applications such as flow control \cite{rabault2019artificial, maceda2021stabilization, deda2023backpropagation, deda2024neural} and shape optimization \cite{zhang2021multi, viquerat2021direct, keramati2022deep, sun2023physics}, the downstream task rarely requires a high-fidelity reconstruction of the entire state vector. Instead, control laws are typically computed from a limited set of localized sensor measurements, while shape optimization relies on cost functions of integrated quantities derived from the flow field. In this context, surrogates that provide full-state reconstructions suffer from two fundamental shortcomings.
First, their training objective, which is typically the minimization of the global reconstruction error across the entire domain, does not directly prioritize the quantities of engineering interest, whether these are pointwise sensor readings or integral cost functionals. Second, the requirement to reconstruct the full-order state imposes higher model complexity, incurring unnecessary computational overhead that hinders performance during both training and, more critically, online deployment where latency constraints are strict. These limitations motivate the development of surrogate models that operate directly in a sparse measurement space, bypassing full-state reconstruction in favor of targeted predictions of task-relevant observables.

%
%

Following this rationale, we employ graph neural networks (GNNs) on coarse graphs whose nodes are strategically placed to encompass sensor locations and regions relevant for the evaluation of integral metrics.
This type of neural network
can efficiently resolve spatial interactions through localized message-passing algorithms by representing the discretized domain as a graph, where nodes represent spatial points and edges encode local connectivity \cite{gilmer2017neural, battaglia2018relational}. Foundational works have demonstrated that GNN-based simulators can capture complex physical dynamics across diverse domains, such as Lagrangian \cite{sanchez2020learning, LANTERI2025106773} and Eulerian-based mechanics \cite{pfaff2021,fortunato2022multiscalemeshgraphnets}. Physics-informed graph neural Galerkin networks have been successfully formulated to solve forward and inverse PDE problems \cite{gao2022physics}, while temporal GNN frameworks constrained by physical flow information have shown excellent capability in simulating granular systems \cite{zhao2025physical}.
Furthermore, solutions of low fidelity simulators have been enhanced by graph neural networks-based correctors, which could aid engineers at early design stages \cite{SHEN2026107153}. While these high-fidelity graph models typically operate on dense, full-order simulation meshes, we propose utilizing GNNs on sparse probe-based graph topologies, establishing an efficient reduced-order modeling pathway.

In this work, we present a parameterized graph-based ROM framework trained to accurately predict the dynamical evolution of unsteady flows over a range of parameters, including unseen operating regimes. Rather than relying on dense computational grids, our GNNs operate directly on coarse meshes constructed from sensor probes at key locations of interest. First,  an ablation study is performed on two canonical two-dimensional flow problems: a cylinder immersed in a plane channel for a range of Reynolds numbers, and an asymmetrically heated cavity flow at different Rayleigh numbers featuring intermittent dynamics. As a final test, we deploy the coarse-mesh GNNs on a three-dimensional transitioning flow over a confined cylinder. Model performance is assessed through autoregressive rollouts at sensor locations and integrated engineering quantities of interest, such as the drag coefficient and Nusselt number. System identification and state estimation tasks are also performed.

The present framework synthesizes three complementary techniques: mesh coarsening to reduce state dimensionality while preserving accuracy, non-uniform parameter-space sampling to concentrate training data in regions of rapidly changing dynamics, and unrolling scheduling during training to mitigate long-term error accumulation. By unifying these strategies, this work yields accurate, computationally efficient graph-based ROMs and delivers three primary contributions. First, we demonstrate that GNNs operating directly on coarse spatial topologies match the predictive accuracy of fine-mesh baselines, even when considering complex 3D systems that require millions of full-order grid points to be simulated, while eliminating full-state reconstruction overhead, thereby lowering memory requirements and training times. Second, our dynamic training pipeline, which combines adaptive parameter sampling with noise-augmented unrolling, suppresses autoregressive error accumulation better than standard linear and non-linear baselines under identical setups. Finally, we establish the capability of these parameterized coarse-mesh surrogates to perform robust system estimation and parameter identification from sparse sensor measurements, reliably predicting flow dynamics across unseen parameter regimes.

\section{Methodology}
\label{sec:methodology}
\subsection{Graph Neural Network Architecture}
\label{sec:gnn}
A graph $\mathcal{G} = (\mathcal{V}, \mathcal{E})$ is defined by a set of $V$ nodes $\bm{v}^i \in \mathcal{V}$ and a set of $E$ edges $\bm{e}^{ij} \in \mathcal{E}$ that connect nodes $\bm{v}^i$ and $\bm{v}^j$. In this work, we consider the flow solution at a discrete time $t$ interpolated to an unstructured mesh to be an undirected graph $\mathcal{G}_t$. Each node at time $t$ $\bm{v}^i_t \in \mathbb{R}^F$ contains $F$ features relevant to the system considered, such as velocity, temperature, pressure, and a one-hot vector used to identify different boundary conditions. Each edge $\bm{e}^{ij}$ contains three values: the differences between the spatial coordinates $\bm{x}^i - \bm{x}^j$ and its Euclidean norm $\norm{\bm{x}^i - \bm{x}^j}$.

The output of the graph neural network model $\mathcal{M}(\mathcal{G}_t)$ is, for each node, a vector $\Delta \bm{v}^i_t$ predicting how the physical input features change from the current time instant to the next, as well as a scalar field ${s}^i_t \in \mathbb{R}$ which acts as a regularizer. More details about input features and scalar fields for the investigated systems in this work are given in Sec.~\ref{sec:systems}. An alternative would be to directly predict the next states, however doing so tends to be more numerically unstable, especially during long rollouts. It is possible then to use the model autoregressively to march the solution in time as

\begin{align}
    & \mathcal{M}(\mathcal{G}_t) = \bm{z}^i_t = \begin{bmatrix} \Delta \bm{v}^i \\ s^i \end{bmatrix}_t = \begin{bmatrix} \Delta v^i_{1} \\ \vdots \\ \Delta v^i_{P} \\ s^i \end{bmatrix}_t \\
    %
    & \bm{v}^i_{t+1} = \bm{v}^{i}_{t} + \Delta \bm{v}^{i}_{t}  \label{eq:gnn_timestep}
\end{align}
where $P < F$ is the number of physical features that evolve over time (excluding fixed features such as one-hot boundary condition encoders). The subscript $t$ denoting time will be ommited for clarity unless necessary.

\begin{figure}[pos=t!]
    \centering
    \includegraphics[width=0.8\textwidth]{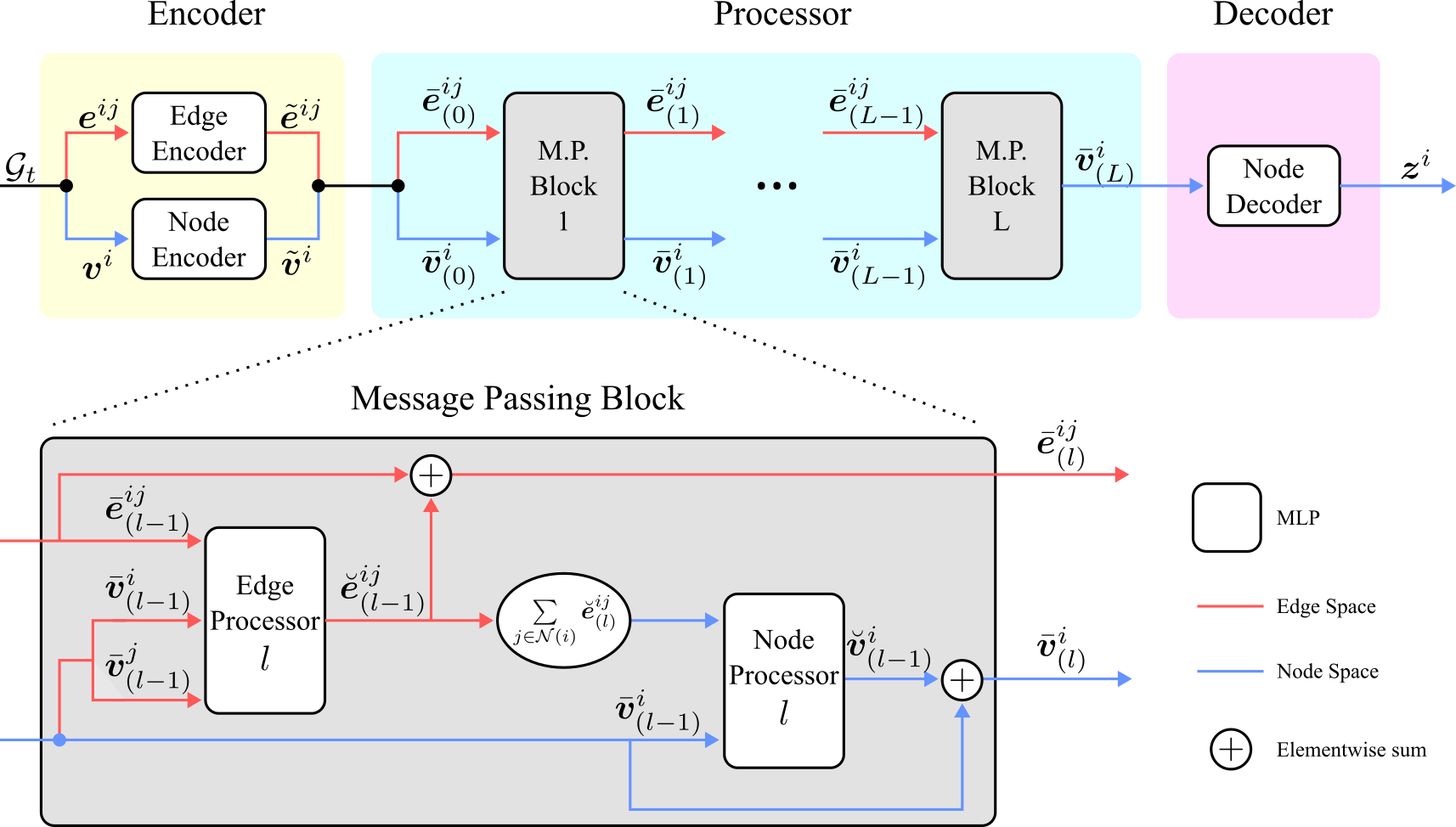}
    \caption{Architecture of GNNs used in this work. Red and blue arrows denote variables in edge (matrices with $E$ rows) and node (matrices with $N$ rows) spaces.}
    \label{fig:architecture}
\end{figure}

We choose a MeshGraphNets-based architecture \cite{pfaff2021}, with a schematic displayed in Fig.~\ref{fig:architecture}, due to its proven ability to accurately predict the solution of PDEs over time and its flexibility regarding mesh discretizations. This architecture works with an Encode-Process-Decode pipeline followed by an integrator. Node encodings $\tilde{\bm{v}}^{i}$ are created by applying a multilayer perceptron (MLP) with the node features as inputs. Edge encodings $\tilde{\bm{e}}^{ij}$ follow the same procedure, but using edge features as inputs
\begin{align}
\tilde{\bm{e}}^{ij} = \text{MLP}(\bm{e}^{ij})&, \\
\tilde{\bm{v}}^{i} = \text{MLP}(\bm{v}^i) \mbox{ .}
\end{align}

The processor is the ensemble of $L$ messaging passing blocks. Each one is defined by two different MLPs that act on edges and nodes, respectively, creating latent variables $\bar{\bm{e}}^{ij}_{(l)}$ and $\bar{\bm{{v}}}^{i}_{(l)}$ where $l \in \{1, \dots, L\}$.
A processor block receives as input the output of the previous one, with the first block taking the encoded node and edge features as inputs. More specifically, the edge processor uses connected nodes and edge encodings as inputs while the node processor uses the node and neighboring edges' encodings aggregate sum as inputs:
\begin{align}
    & \breve{\bm{e}}^{ij}_{(l-1)}  = \text{MLP}_{(l)}([\bar{\bm{{e}}}^{ij}_{(l-1)}, \bar{\bm{{v}}}^{i}_{(l-1)}, \bar{\bm{{v}}}^{j}_{(l-1)}]) \mbox{ ,}\\
    & \breve{\bm{v}}_{(l-1)}^{i} =  \text{MLP}_{(l)}([\bar{\bm{v}}_{(l-1)}^{i}, \textstyle{\sum_{j \in \mathcal{N}(i)}} \, \breve{\bm{e}}^{ij}_{(l)}] ) \mbox{ ,}\\
    & \bar{\bm{e}}_{(l)}^{ij} = \bar{\bm{e}}_{(l-1)}^{ij} + \breve{\bm{e}}_{(l-1)}^{ij} \mbox{ ,}\\
    & \bar{\bm{v}}_{(l)}^{i} = \bar{\bm{v}}_{(l-1)}^{i} + \breve{\bm{v}}_{(l-1)}^{i}\mbox{ ,}
\end{align}
where $\mathcal{N}(i)$ in the node processor MLP is the set of  nodes connected to node $i$.
%
In practice, increasing the number of message-passing blocks extends the effective receptive field (the maximum extent of the graph that influences a single node's embedding) of each node, enabling information to propagate over longer spatial distances within the graph. Although each update step is locally constrained to direct neighbors, the iterative nature of the scheme implies that information from distant nodes is progressively incorporated as messages are transmitted along intermediate vertices via local updates. Thus, the number of message passing blocks directly governs the length of this information propagation path, and consequently, the degree of non-local coupling captured by the model. Skip connections applied to the edge and node processors enable the use of many message-passing blocks while maintaining efficient gradient flow. Unless stated otherwise, in this work we utilize $L=15$.

Final features are calculated with the decoder defined by a MLP that receives the node encoding from the final processor block, $\bm{z}^i = \text{MLP}(\bar{\bm{v}}_{(L)}^{i}).$
%
All MLPs are comprised of two dense layers and only the first one utilizes an activation function, which we choose to be a ReLU. Also, layer normalization is utilized as a final step in order to stabilize and accelerate training, except for the MLP used by the decoder. The number of latent space features after the encoders is $128$.

\subsection{Dynamical Systems}
\label{sec:systems}

Two fluid systems are used to perform an ablation study of the proposed techniques: a confined cylinder flow and a rectangular cavity with convective heat transfer, both solved in two dimensions. The former has become a common benchmark for evaluating machine learning techniques in fluid systems. The latter poses a more complex problem due to the broader spectrum of spatial and temporal scales. As a final test case, we augment the confined cylinder system by solving it in three dimensions in a transitional regime.

\begin{figure}[pos=ht!]
    \centering
    \begin{subfigure}[t]{0.87\textwidth}
        \centering
        \includegraphics[width=\textwidth]{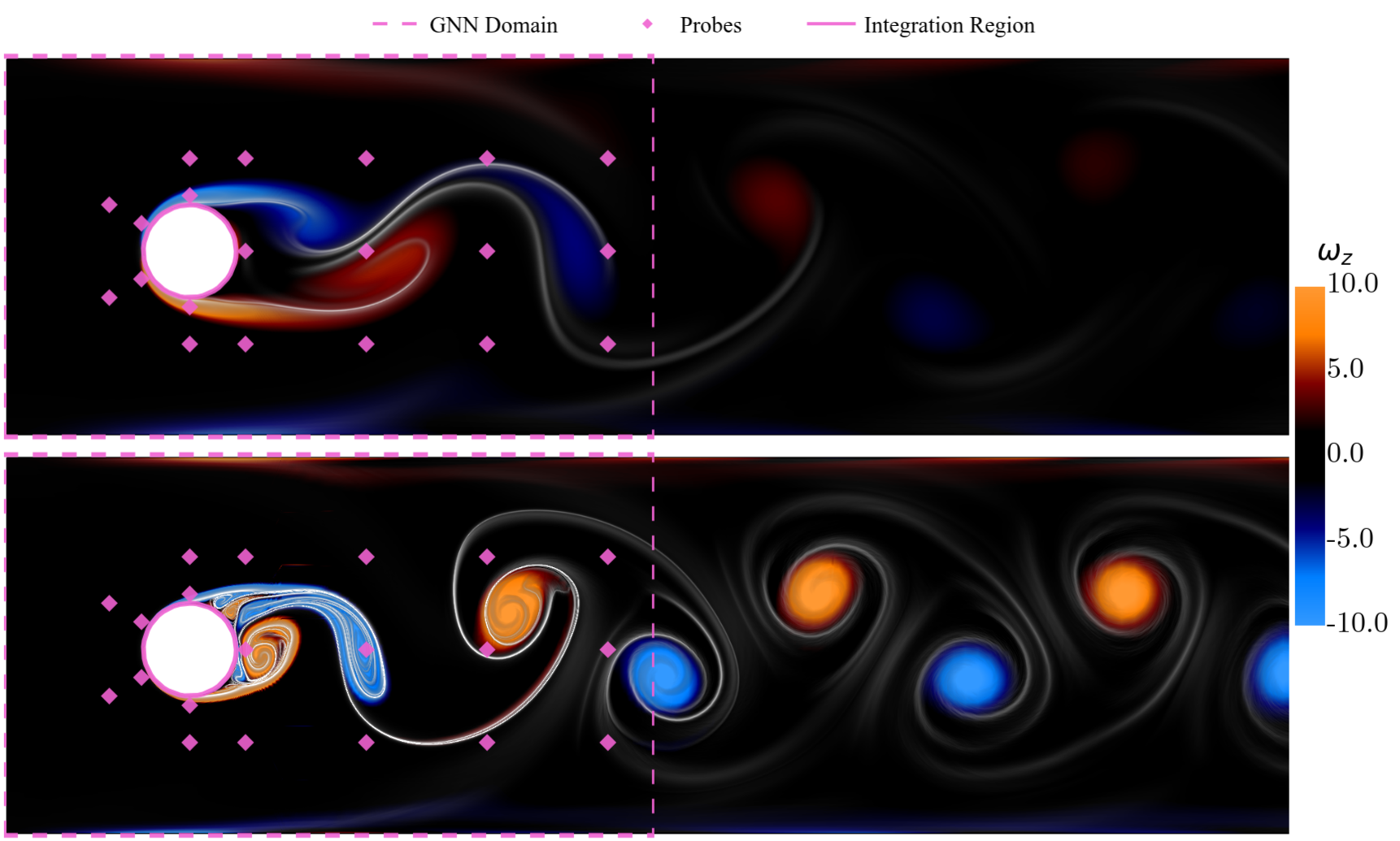}
    \end{subfigure}
    \caption{Contours of vorticity for the 2D confined cylinder flow at $\Rey=100$ (top) and $\Rey=1000$ (bottom). Contours in white show attracting Lagrangian coherent structures through the ridges of backward finite time Lyapunov exponents. Location of probes and integration regions used for evaluating models are highlighted in pink.}
    \label{fig:flows_cyl}
\end{figure}

\begin{figure}[pos=ht!]
    \centering
    \begin{subfigure}[t]{0.75\textwidth}
        \centering
        \includegraphics[width=\textwidth]{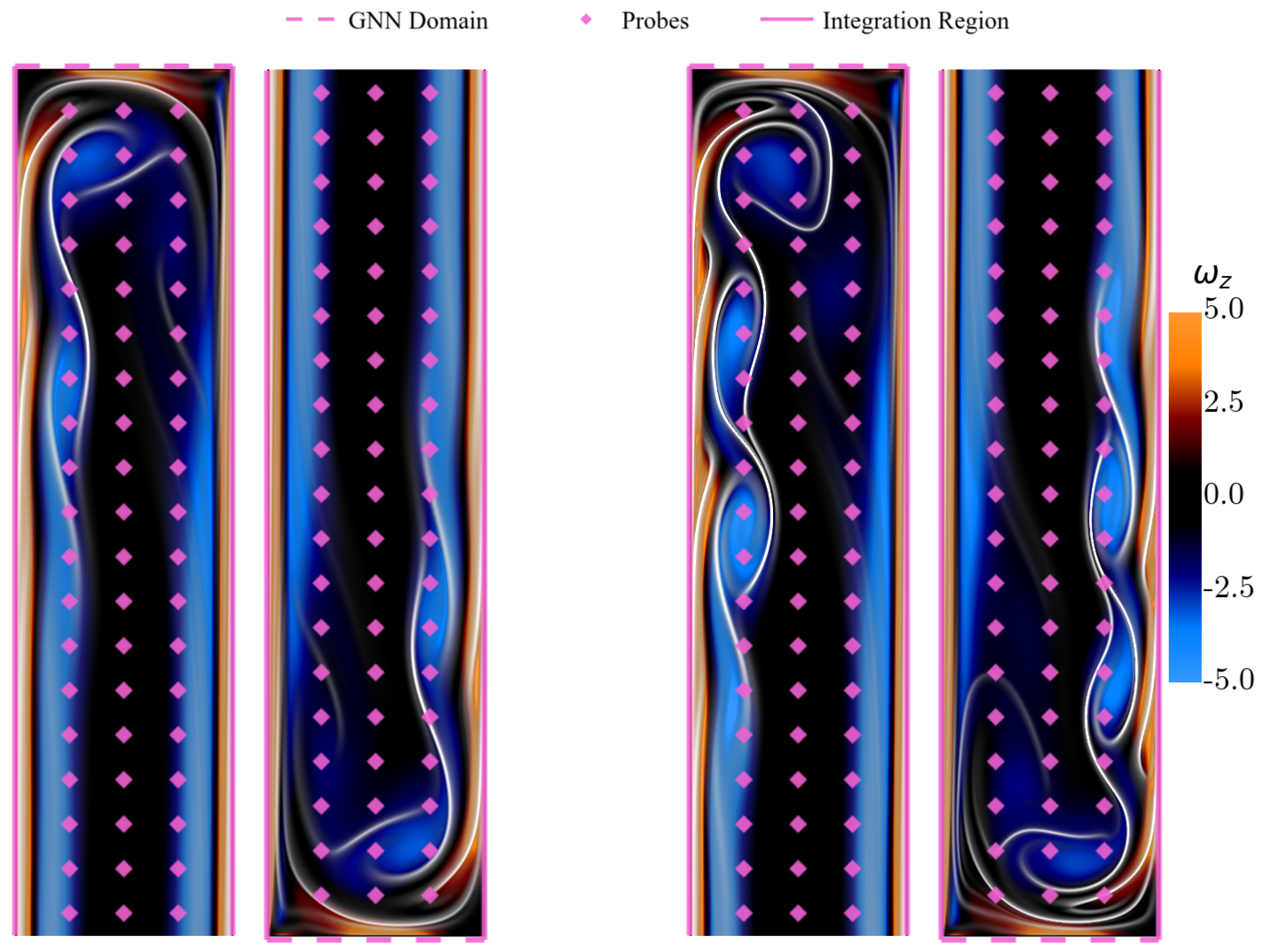}
    \end{subfigure}
    \caption{Contours of vorticity for the cavity flow at $\Ray=4 \times 10^5$ (left) and $\Ray=6 \times 10^5$ (right). For each $\Ray$, the top and bottom portions of the cavity are shown separately for better visualization. Contours in white show attracting Lagrangian coherent structures through the ridges of backward finite time Lyapunov exponents. Location of probes and integration regions used for evaluating models are highlighted in pink.}
    \label{fig:flows_cav}
\end{figure}

\begin{figure}[pos=ht!]
    \centering
    \begin{subfigure}[t]{0.7\textwidth}
        \centering
        \includegraphics[width=\textwidth]{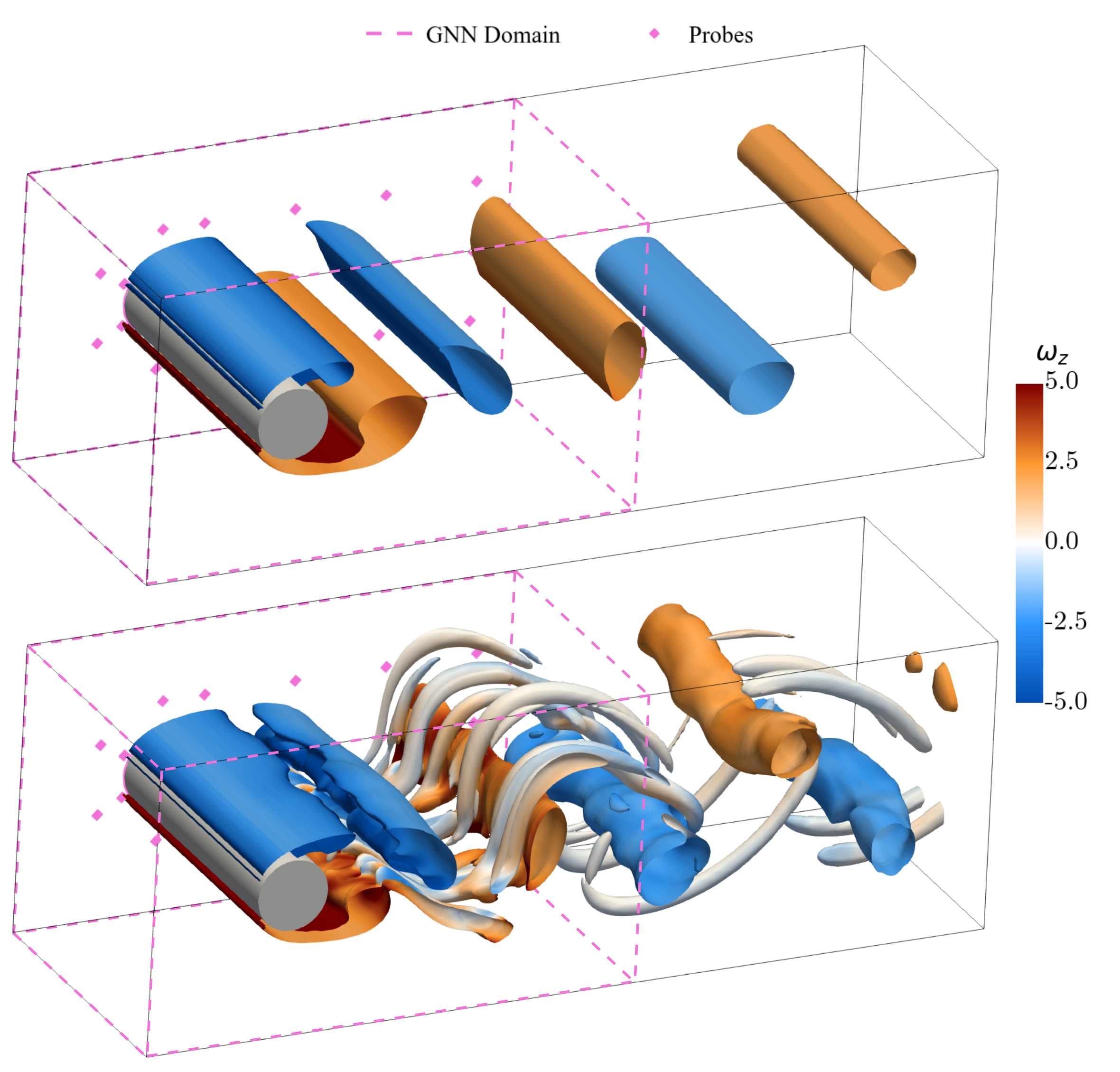}
    \end{subfigure}
    \caption{Q-criterion isosurfaces with contours of vorticity for the 3D confined cylinder system at $\Rey=100$ (top) and $\Rey=200$ (bottom). Location of probes used for evaluating models are highlighted in pink. The whole cylinder surface is the integration surface for this configuration.}
    \label{fig:flows_cyl3D}
\end{figure}

All system dynamics are dictated by a non-dimensional parameter: the Reynolds number $\Rey$ for the cylinder flow and the Rayleigh number $\Ray$ for the cavity case. Increasing these parameters drives the systems to more complex trajectories, thereby making prediction tasks fundamentally harder. This behavior can be seen in Figures~\ref{fig:flows_cyl}, \ref{fig:flows_cav} and \ref{fig:flows_cyl3D}.
For reduced-order models to be practically useful, they must perform reliably across a range of parameter values, not only at a fixed setting. Accordingly, we generate datasets from both systems spanning different non-dimensional flow parameters. Details about p-refinement studies to ensure that all simulations are properly converged for the investigated parameter space are shown in the Appendix \ref{app:convergence}.
All simulations are performed using the Nek5000 spectral element code \cite{fischer2008} coupled with the KTH Framework \cite{massaro2024}.

\subsubsection{Confined Cylinder (2D)}

The two-dimensional confined cylinder case is characterized by the flow over a cylinder placed inside a plane channel, where the cylinder is slightly offset from the center of the channel. This particular configuration has been selected as it represents a canonical test case commonly used for flow prediction and control \cite{ren2021, mao2022, zhang2022, deda2024neural}. 
We numerically solve the incompressible Navier-Stokes equations in non-dimensional form defined by
\begin{align}
    & \nabla \cdot \bm{u} = 0 \mbox{,} \label{eq:inc}\\
    & \frac{\partial \bm{u}}{\partial t}  = - \bm{u} \cdot \nabla \bm{u} - \nabla p +  \frac{1}{\Rey}\nabla^2\bm{u} \label{eq:ns} \mbox{ ,}
\end{align}
where $\bm{u} = [ u_x \; u_y ]^T$ is the velocity vector, $p$ is the pressure, 
and $\Rey$ is the Reynolds number.
No-slip boundary conditions are imposed on the cylinder surface as well as in the upper and bottom channel walls. The inlet boundary condition is set as a parabolic profile following ${u}_x(x\!=\!0, y) = 6 (H/2 - y) (H/2 + y)/H^2$, and $u_y(x\!=\!0, y)=0$, where $H=4.1$ is the channel height and the cylinder diameter is unitary. Finally, the outlet boundary condition is implemented as a Neumann condition. In some analysis, we use pressure drag which is calculated by numerically integrating pressure at the cylinder wall according to
\begin{align}
    C_D = \int_\text{cyl wall} p \hat{n} \cdot \hat{i} \, \textit{dA} \mbox{ .}
\end{align}
where $\hat{n}$ is the outward normal with respect to the wall and $\hat{i}$ is a unitary horizontal vector.

\begin{figure}[pos=ht!]
    \centering
    \begin{subfigure}[b]{0.35\textwidth}
        \centering
        \includegraphics[height=\textwidth]{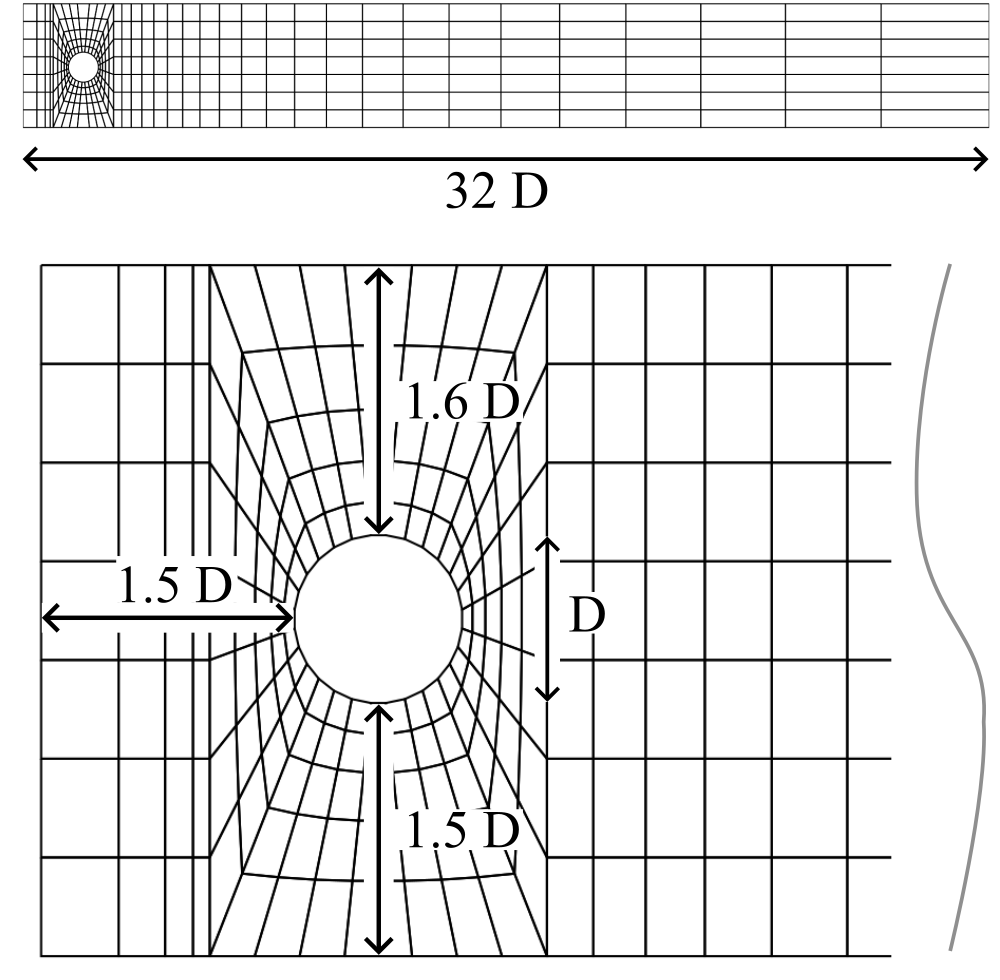}
    \end{subfigure}
    \begin{subfigure}[b]{0.35\textwidth}
        \centering
        \includegraphics[height=\textwidth]{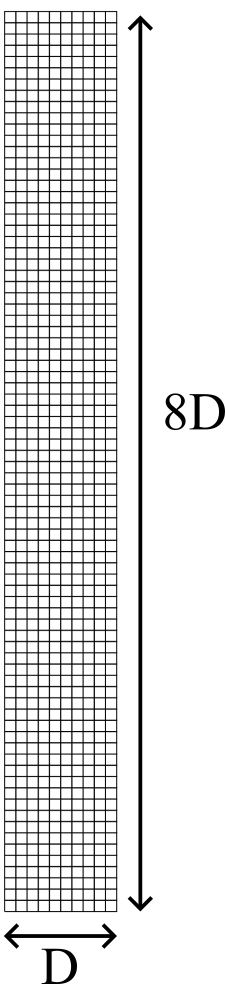}
    \end{subfigure}
    \caption{Simulation domains for the confined cylinder (left) and cavity  (right) systems, respectively, where quadrilaterals 
    represent spectral elements.}
    \label{fig:sim_domain}

\end{figure}

The inlet and outlet are situated $2$ and $30$ spatial units away from the cylinder center, respectively. There are a total of $322$ spectral elements, which totals $32\,200$ or $63\,112$ points, depending on the polynomial order used within the elements. They are concentrated near the cylinder and stretched as approaching the outlet, where a sponge is used to avoid numerical instabilities. More details about the domain size are shown in Fig.~\ref{fig:sim_domain}.

\begin{figure}[pos=!ht]
    \centering
    \begin{subfigure}[t]{0.48\textwidth}
        \centering
        \includegraphics[width=\textwidth]{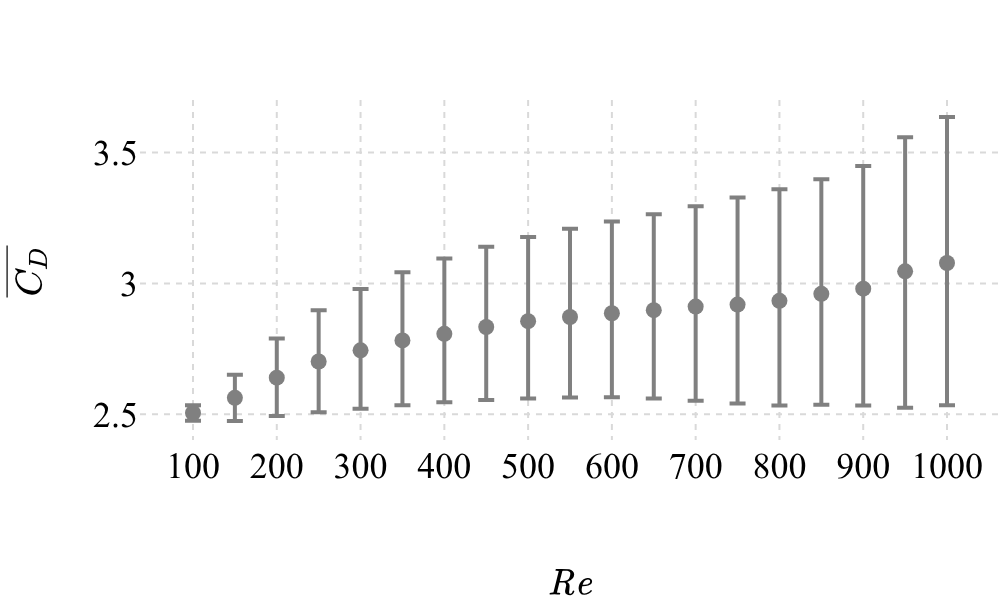}
        \caption{Mean $C_D$ where bars represent maximum and minimum values.}
        \label{fig:cyl_dynamicsa}

    \end{subfigure}
    \begin{subfigure}[t]{0.48\textwidth}
        \centering
        \includegraphics[width=\textwidth]{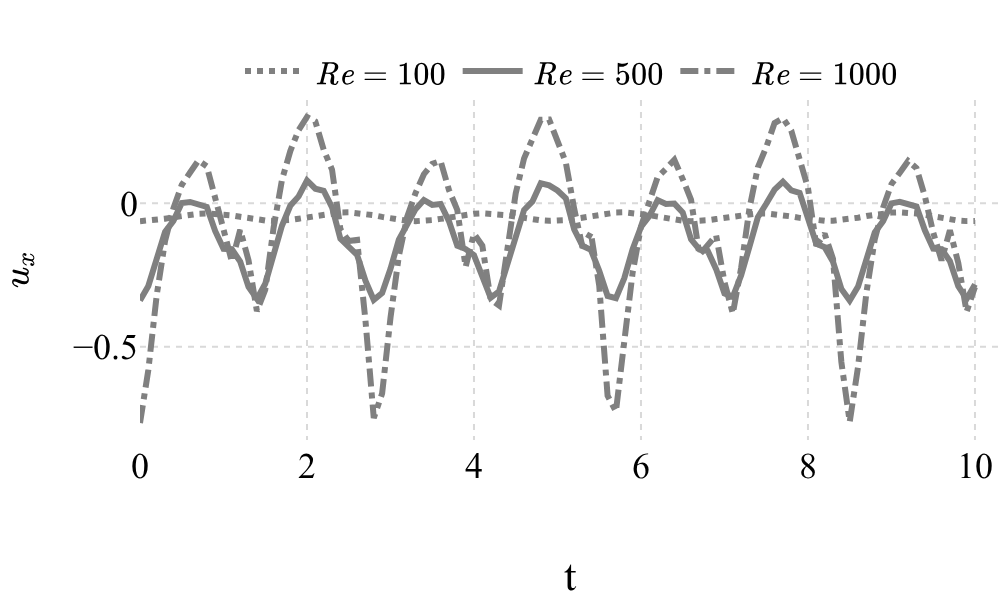}
        \caption{Horizontal velocity signal taken at $x=0.6, y=-0.05$.}
        \label{fig:cyl_dynamicsb}
    \end{subfigure}
    \caption{2D cylinder dynamics at different $\Rey$.}
    \label{fig:cyl_dynamics}
\end{figure}

The parameter space for this system is $\Rey \in [100, 1000]$. This range was chosen because it exhibits vortex shedding with a different range of spatial scales and amplitudes, as shown in Fig.~\ref{fig:flows_cyl}.
Variations in mean drag coefficient, as well as maximum and minimum values, can be observed in Fig. \ref{fig:cyl_dynamicsa}. At $\Rey = 100$, drag oscillations are minimal but grow rapidly as $\Rey$ increases up to approximately $\Rey = 300$, after which the changes become less pronounced. However, near $\Rey = 900$ the dynamics begin to change rapidly once again. Signals of a probe near the trailing edge are presented in Fig. \ref{fig:cyl_dynamicsb}, showing how increasing the $\Rey$ number affects the frequency components present in the system and their amplitudes.

Datasets used for training are created from $10$ simulations with different $\Rey$, with data collected after initial transients have decayed. The non-dimensionalized timestep used in the simulations is $\Delta t_\text{cyl}^* = 0.001$, with the flow field being saved every 100 timesteps. This implies that the model timestep is $\Delta t_{\text{cyl}} = 0.001 \times 100 = 0.1$, a considerably larger time step compared to the one used by the numerical solver. Each simulation is executed until 4000 snapshots are exported, with $90\%$ used for training and $10\%$ for validation. After training, the model is deployed autoregressively as described in Eq. \ref{eq:gnn_timestep} on initial conditions not used during training, for a range of $\Rey$ values starting from 100 to 1000, increasing in increments of 50. A total of $400$ timesteps, which are equivalent to $40$ convective time units, are run during the test phase for each Reynolds number.

For this system there are $F_\text{cyl} = 7$ node features: the velocity components $u_x^i$ and $u_y^i$, Reynolds number $\Rey$ and a one-hot vector describing if that node is an inlet, an outlet, a wall or a fluid. Predictions are the velocity variations $\Delta u_x^i$ and $\Delta u_y^i$ ($P_\text{cyl} = 2$), as well as the pressure at each node $s_\text{cyl}^i = p^i$.


\subsubsection{Cavity with Convective Heat Transfer}

The other system used in this work refers to a cavity flow driven by buoyancy forces due to a temperature difference at the walls. More specifically, fluid is heated from the left and cooled from the right-hand side, with the top and bottom walls being adiabatic. The presence of a gravitational force in the negative vertical direction induces a buoyant force which causes hot and cold fluid to move upward and downward, respectively. Despite its simple geometry, this setup offers increasingly rich dynamics as the thermal forcing is increased \cite{Tkachenko01102016,Zucatti2020ROM}, thus making it a suitable candidate to evaluate the performance of surrogate models.

Simulations are carried out assuming incompressibility, where continuity is also defined by Eq.~\ref{eq:inc}. The momentum and energy equations take the form
\begin{align}
    & \frac{\partial \bm{u}}{\partial t}  = - \bm{u} \cdot \nabla \bm{u} - \nabla p + \sqrt{\frac{\Prn}{\Ray}}\nabla^2\bm{u} + T\bm{\hat{j}} \label{eq:ns+bou} \mbox{ , and} \\
    &\frac{\partial T}{\partial t} + \bm{u} \cdot T = \frac{1}{\sqrt{\Prn \Ray}} \nabla^2T \mbox{ ,}
\end{align}
where $T$ represents the temperature,
$\Ray$
is the Rayleigh number, and
$\Prn=0.71$ is the Prandtl number, with its value representative of air. A buoyancy term in Eq.~\ref{eq:ns+bou} emerges from the Boussinesq approximation, which creates the forcing that modifies the flow. No-slip boundary conditions are imposed on all walls. Regarding temperature, the left and right side walls have a temperature difference $\Delta T = T_{x=0} - T_{x=W} = 1 - 0 = 1$.

Snapshots of the cavity flows at different Rayleigh numbers are presented in Fig.~\ref{fig:flows_cav}, where buoyancy-driven coherent structures are advected along the top and bottom walls of the cavity.
The cavity has a unitary width length $D$ and aspect ratio of $1\text{:}8$ as displayed in Fig.~\ref{fig:sim_domain}. A total of $800$ spectral elements are distributed uniformly throughout the domain, totaling $80\,000$ or $156\,800$ points depending on the polynomial order utilized.

\begin{figure}[pos=!ht]
    \centering
    \begin{subfigure}[t]{0.48\textwidth}
        \centering
        \includegraphics[width=\textwidth]{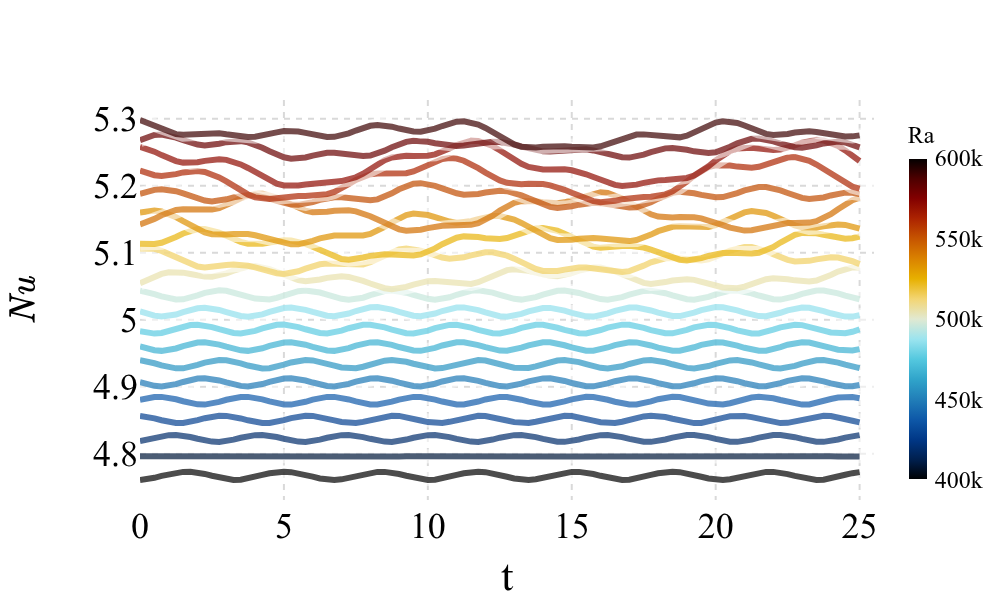}
        \caption{$\Nus$ variations for various $\Ray$ across time.}
    \end{subfigure}
    \begin{subfigure}[t]{0.48\textwidth}
        \centering
        \includegraphics[width=\textwidth]{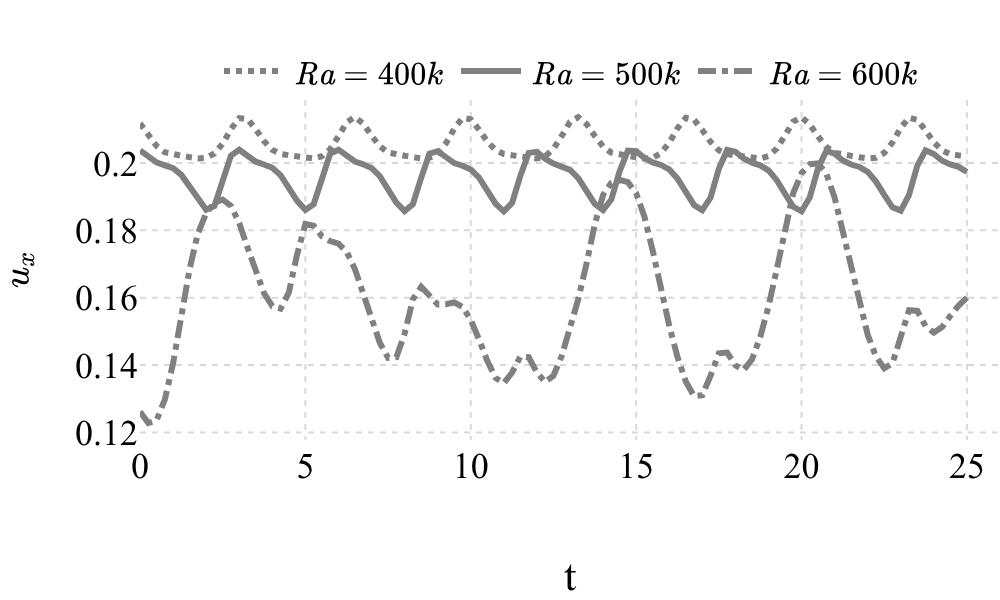}
        \caption{Horizontal velocity signals taken at $x=0.25, y=7.79$ for different $\Ray$.}
    \end{subfigure}
    \caption{Cavity flow dynamics according to $\Ray$.}
    \label{fig:cav_dynamics}
\end{figure}

Simulations are performed over the range $\Ray \in [4\times 10^5, 6\times 10^5]$, with the flow variations shown as a function of time in Fig.~\ref{fig:cav_dynamics} for both the Nusselt number $\Nus$ and the horizontal velocity component $u_x$. For visual clarity, figures use the shorthand notation where ``$\mathrm{k}$'' denotes $10^3$ (e.g. $400\mathrm{k}$ for $4 \times 10^5$).
In the first half of the parameter space, the dynamics are purely periodic and exhibit a clear dominant frequency. However, at $\Ray = 4.1\times 10^5$, the temperature gradients at the wall become out of phase, which effectively renders $\Nus$ constant, an interesting dynamical shift. In the second half of the parameter space, the frequency spectrum becomes richer, and by the upper end the flow is no longer periodic. Also, the amplitude of $\Nus$ does not increase monotonically with $\Ray$. Thus, this system exhibits more challenging dynamics when compared to the cylinder case, where the flow stays periodic throughout the whole parameter space investigated in this work.

The training dataset for this system is composed of $11$ simulations. They are performed with a characteristic timestep of $\Delta t^*_{\text{cav}} = 0.005$ with snapshots being saved every $50$ timesteps, implying an effective timestep of $\Delta t_\text{cav} = 0.25$ for the surrogate model. A total of $3600$ snapshots are gathered for each simulation after initial transients have passed and, as with the cylinder case, $90\%$ of the data is used for training and $10\%$ for validation. The test simulations contain $400$ snapshots, which represent $100$ convective time units. They are performed with $\Ray$ numbers starting from $4\times 10^5$ to $6\times 10^5$ with increments of $10^4$.

Graphs of this system contain $F_\text{cav} = 7$ node features: the velocity components $u_x^i$ and $u_y^i$, temperature $T^i$, Rayleigh number $\Ray$ and a one-hot vector indicating if a node is a wall, temperature precribed or adiabatic, or fluid. Predictions are sought for the velocity and temperature variations $\Delta u_x^i$, $\Delta u_y^i$ and $\Delta T^i$ ($P_\text{cav} = 3$). The temperature gradient in the horizontal direction $s_\text{cav}^i = \nabla_x T^i$ is also predicted at each node, which is used to calculate the Nusselt number as
\begin{align}
    \Nus = \frac{1}{2H} \frac{D}{\Delta T} \left( \int_0^H \left. \frac{\partial T}{\partial x}\right|_{x=0} dy + \int_0^H \left. \frac{\partial T}{\partial x}\right|_{x=D} dy \right) \mbox{.}
\end{align}


\subsubsection{Confined Cylinder (3D)}
As a final test case, we use the same configuration as the two-dimensional cylinder flow but extended in the spanwise direction. This considerably increases the complexity of the dynamical system through the three-dimensional breakdown of vortices shed from the cylinder.

The Navier--Stokes equations are solved according to Eq.~\ref{eq:ns}, with $\bm{u} = [u_x\; u_y\; u_z]^T$. Periodic boundary conditions are imposed in the spanwise direction, in addition to those used for the 2D configuration. 

The mesh for this configuration is generated by extruding the mesh from the 2D confined cylinder case until $z_{\text{max}}=4.1$ using 7 spectral elements in the spanwise direction. The resulting mesh contains $2\,254\,000$ points distributed across $2254$ elements.

\begin{figure}[pos=!ht]
    \centering
    \begin{subfigure}[t]{0.48\textwidth}
        \centering
        \includegraphics[width=\textwidth]{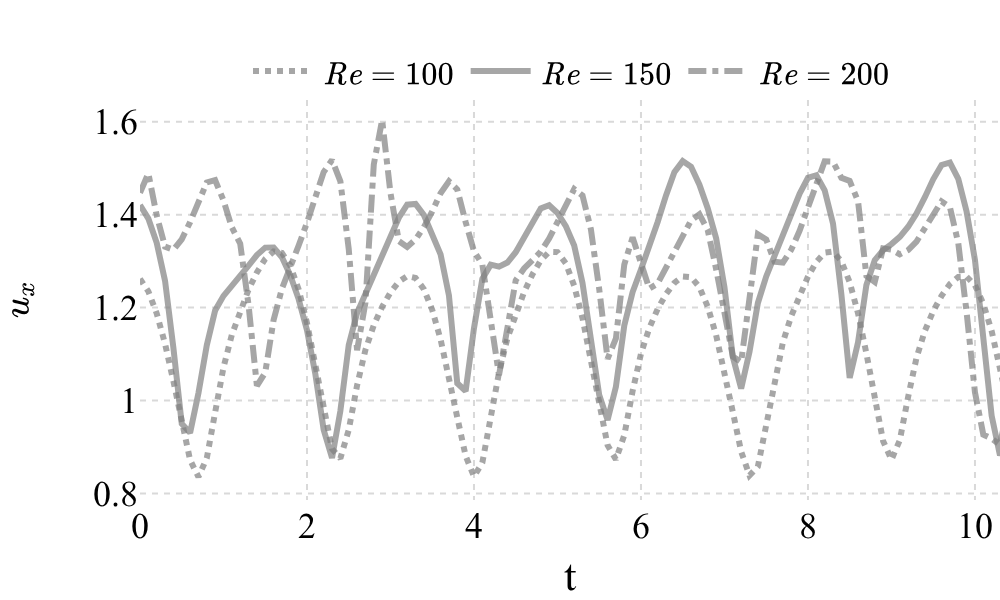}
    \end{subfigure}
    \begin{subfigure}[t]{0.48\textwidth}
        \centering
        \includegraphics[width=\textwidth]{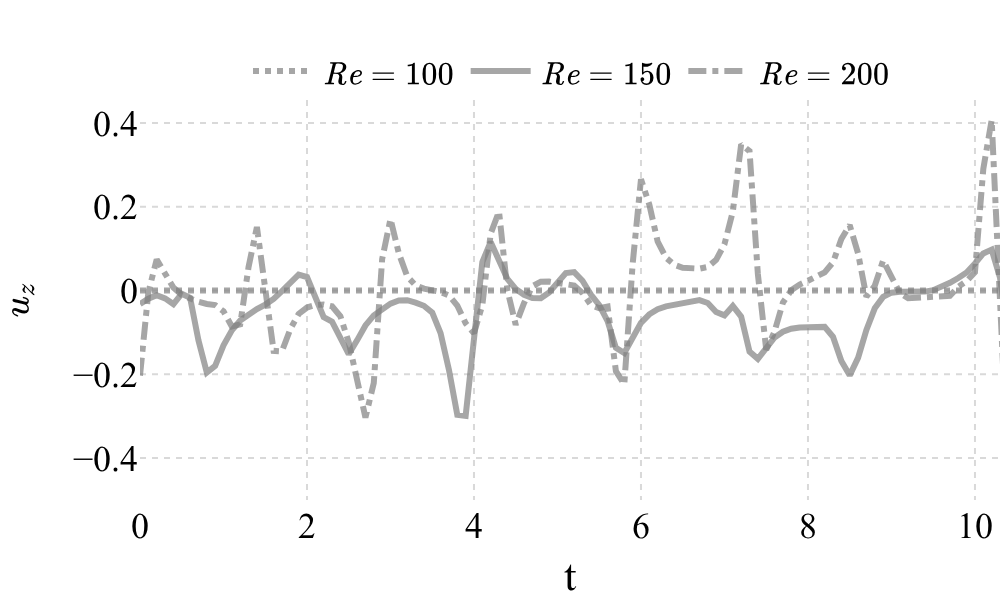}
    \end{subfigure}
    \caption{3D cylinder dynamics at different $\Rey$. Horizontal (left) and and spanwise (right) velocity signals taken at $x=3.2, y=-0.05$.}
    \label{fig:cyl3D_dynamics}
\end{figure}

The Reynolds number range considered is $\Rey \in [100, 200]$. This is a transitional regime where at $\Rey=100$ the flow is completely two-dimensional and time-periodic, whereas for $\Rey\ge150$ three-dimensional intermittent effects are present, increasing the complexity of the system dynamics. The models must therefore learn very different temporal evolutions depending on the Reynolds number, as shown in Fig.~\ref{fig:cyl3D_dynamics}.

Training is performed using datasets generated from 3 simulations, each run until 2800 snapshots have been exported, where $90\%$ of it is used for training and $10\%$ for validation. Testing is conducted over a range of $\Rey$ values from 100 to 200 in increments of 25. During the test phase, 150 timesteps are used for each Reynolds number, which is equivalent to 15 convective time units.

This system uses node features similar to those of the 2D configuration but includes the spanwise velocity $u_z^i$, thus $F_\text{cyl3D} = 8$. Predictions are the velocity variations in all three directions ($P_\text{cyl3D} = 3$). All parameters not explicitly mentioned are kept the same as in the 2D case.


\subsection{Unstructured Mesh Design}

It is common to utilize the same mesh used by the original solver when training surrogates based on graph neural networks. Numerical solvers tend to converge to an accurate solution as mesh resolution increases but there is no guarantee this is also the case for GNNs. This motivates a mesh refinement study, where models are trained on graphs with different levels of refinement to understand their impact on the final accuracy.

The unstructured meshes are generated with two primary considerations. First, they are constrained to include a specific set of points that serve as evaluation probes. Although some engineering applications require full knowledge of the flow domain, many rely on reduced-order representations of the system, such as those used in control-oriented modeling. Therefore, models are tested regarding their capacity for accurately predicting flow properties at specific probe locations.
Second, sufficient refinement should exist at particular regions  to accurately compute flow integrated quantities such as $C_D$ and $\Nus$. To maintain consistency across different resolutions, all meshes, regardless of coarseness or fineness, are constructed with an identical number of points in those regions where surface integration is performed, thereby ensuring that comparisons of integrated variables are not biased by discretization differences. Figures~\ref{fig:flows_cyl}, \ref{fig:flows_cav} and~\ref{fig:flows_cyl3D} show the exact locations of evaluation probes and integration regions. In total, the cylinder and cavity problems employ $B=20$ and $B=114$ probes, respectively. 
For the 3D cylinder case, probes are distributed along a single 2D slice of the domain as shown in Fig.~\ref{fig:flows_cyl3D}.




The number of nodes and edges of each unstructured mesh are compiled in Table~\ref{tab:meshes}. Note that even the finer meshes are considerably coarser than those employed by the numerical solver.
Therefore, the GNNs in this work act as reduced order models when compared to the original solver due to the much larger timestep as well as lower spatial resolution.
For the 3D cylinder case, graphs are generated by extruding 2D unstructured meshes and creating edges between nodes in the first and last slice considering the spanwise-periodic direction. Additional details about the unstructured meshes can be found in the Appendix \ref{app:unstructured_meshes}.

\begin{table}[width=.9\textwidth,cols=4,pos=h]
\caption{Unstructured mesh attributes.}
\label{tab:meshes}
\begin{tabular*}{\tblwidth}{@{} LLLL@{} }
\toprule
System & Name & \# Nodes & \# Edges     \\
\midrule
Cylinder (2D) & C1    & 108      & 240     \\
Cylinder (2D) & C2    & 239      & 611     \\
Cylinder (2D) & C3    & 461      & 1259    \\
Cylinder (2D) & C4    & 978      & 2782    \\
Cylinder (2D) & C5    & 2083     & 6061    \\
Cavity & K1    & 322      & 755     \\
Cavity & K2    & 532      & 1377    \\
Cavity & K3    & 1096     & 3065    \\
Cavity & K4    & 2127     & 6148    \\
Cavity & K5    & 4417     & 13\,004   \\
Cylinder (3D) & T1    & 3629     & 12\,673    \\
Cylinder (3D) & T2    & 14\,906     & 56\,202    \\
Cylinder (3D) & T3    & 29\,261    & 113\,071    \\
\bottomrule
\end{tabular*}
\end{table}


\subsection{Adaptive Parameter Sampling}
\label{sec:methodology_nud}
Acquiring data for training models is often an expensive task, making it possible to sample the parameter space only for a limited amount of realizations.
%
For complex nonlinear systems, the dynamics do not change linearly with parameters such as $\Rey$ and $\Ray$. Therefore, building a training dataset by uniformly sampling the parameter space is likely not an efficient strategy. To address this, we propose a parameter space separation score $\phi(\alpha_A, \alpha_B)$ which measures how different the dynamics are in the parameter space regions $\alpha_A$ and $\alpha_B$. This allows us to identify regions where dynamics shift rapidly and build a training dataset that is non-uniformly sampled.

To operationalize this separation score, we proceed as follows. First, we gather signals at locations of interest (probe points and integration regions). For each signal, we compute the power spectral density (PSD) and normalize it by its total integral. We then compute the Earth Mover's Distance (EMD)~\cite{emd} between PSDs at the same spatial locations and average this distance across all signals. Hence, parameter space regions with similar distributions of signal energy across frequencies are considered to have similar dynamics. For the cylinder and cavity cases, we use $p$ and $\nabla_x T$ as the signals, respectively, with $500$ snapshots per case. This process is described by
\begin{equation}
\phi(\alpha_A, \alpha_B) = \frac{1}{S} \sum_{l=1}^{S} \text{EMD}\left( \text{PSD}\bigl(x_l(\alpha_A)\bigr),\, \text{PSD}\bigl(x_l(\alpha_B)\bigr) \right),
\label{eq:phi}
\end{equation}
where $S$ is the total number of signal locations (probe points and integration regions), $x_l(\alpha)$ is the time-series signal at location $l$ for parameter $\alpha$, and $\text{PSD}(\cdot)$ denotes the power spectral density normalized by its total integral.

We construct the non-uniform training dataset iteratively. First, we initialize the parameter set with the minimum, midpoint and maximum values of the parameter range.  We then compute the separation score $\phi$ between each pair of consecutive samples. The pair with the largest $\phi$ indicates where the dynamics change most rapidly between neighboring parameter regions. Between this pair, we add a new sample at its midpoint. We then recompute $\phi$ for the new consecutive pairs and repeat the process until the desired number of samples is reached.

\begin{figure}[pos=!ht]
    \centering
    \includegraphics[width=0.5\textwidth]{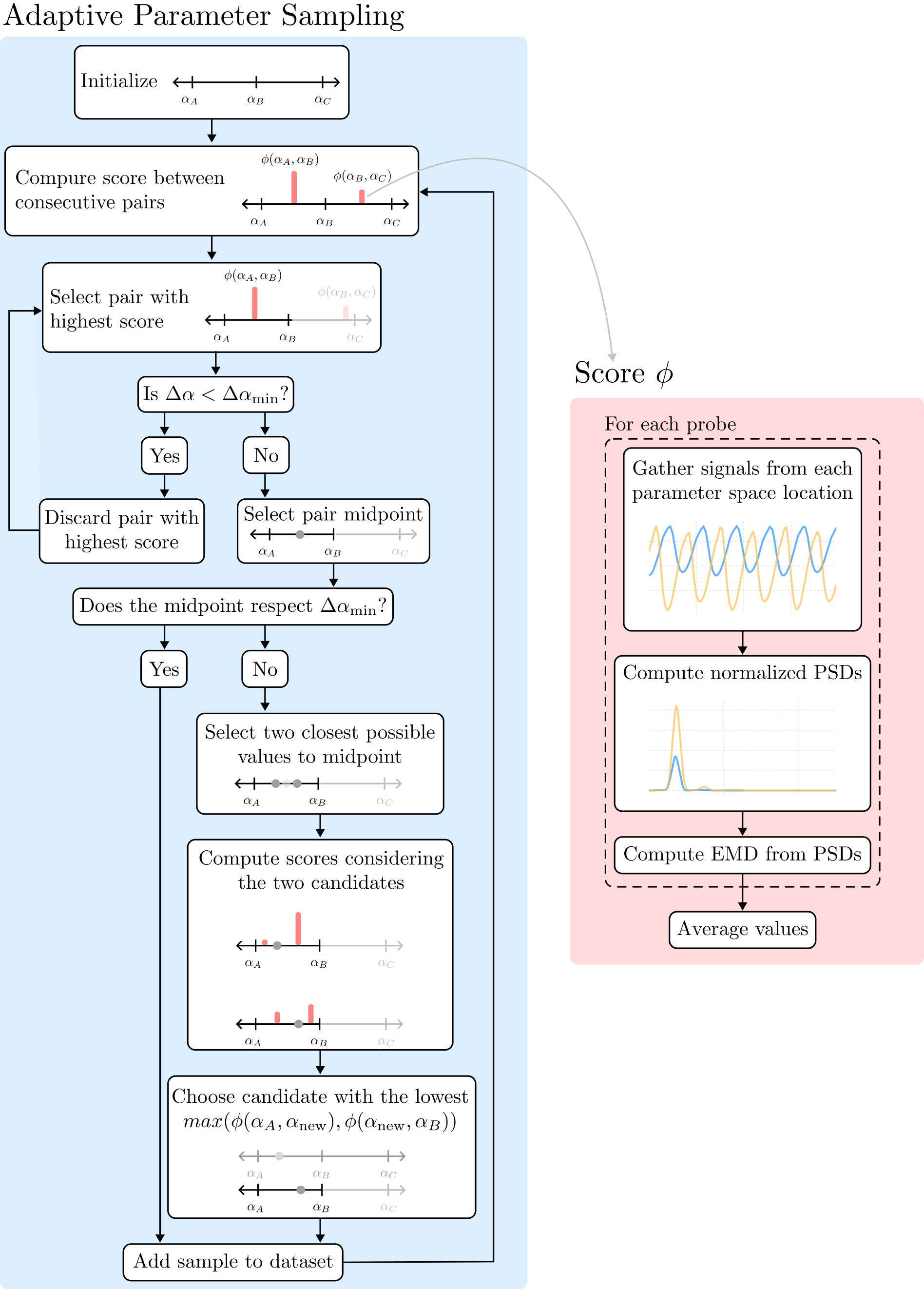}
    \caption{Adaptive parameter sampling procedure.}
    \label{fig:nud_schematic}
\end{figure}

\begin{figure}[pos=!ht]
    \centering
        \includegraphics[width=\textwidth]{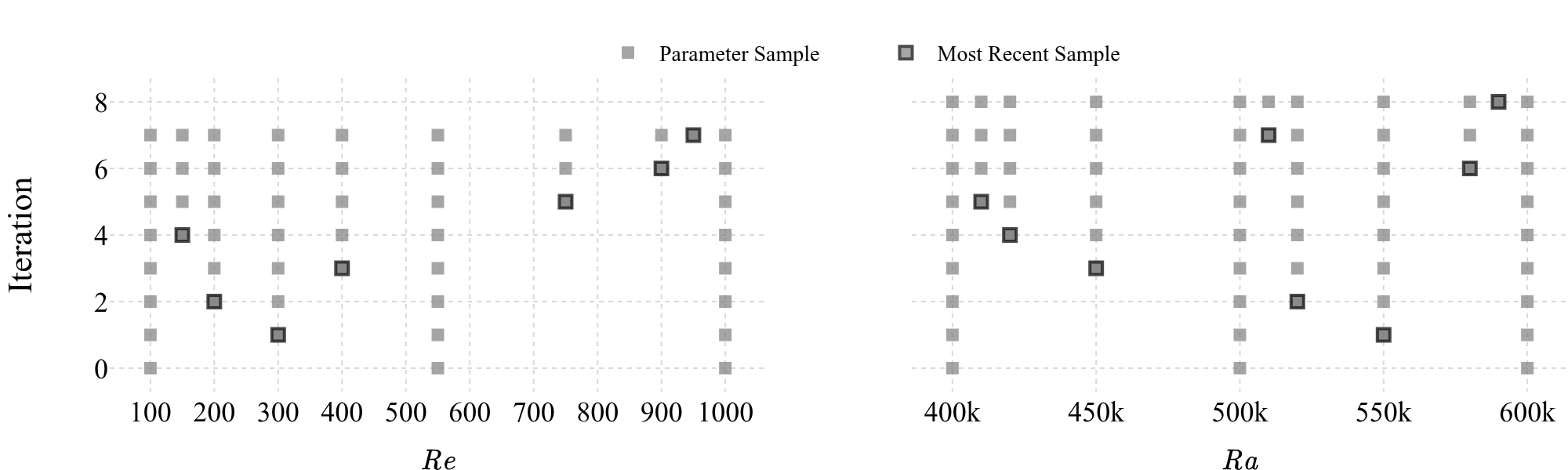}
        \caption{Iterative addition of parameter samples using the proposed adaptive sampling algorithm.}
    \label{fig:nud_iterations}
\end{figure}

In practice, a minimum sampling spacing $\Delta \alpha_{\min}$ exists.
If the midpoint between a consecutive pair falls within $\Delta \alpha_{\min}$ of either existing sample, that pair cannot be further subdivided. Similarly, when adding a sample at the midpoint of a pair $(\alpha_A, \alpha_B)$, we must ensure the new sample also respects the spacing constraint with respect to its neighbors. For candidate locations other than the midpoint, we evaluate the maximum of the two new separation scores it would create, i.e., $\max(\phi(\alpha_A, \alpha_{\text{new}}), \phi(\alpha_{\text{new}}, \alpha_B))$, and select the candidate that minimizes this value. If no permissible location exists for the pair with the largest $\phi$, we skip it and proceed to the pair with the next largest $\phi$. A schematic of the algorithm is shown in Fig.~\ref{fig:nud_schematic} and its result for both systems considered in this work can be seen in Fig.~\ref{fig:nud_iterations}.
Note that this process is model-agnostic and purely system-dependent. Thus, it can be performed prior to any surrogate training.
%

\subsection{Training Strategy}

During long rollouts, autoregressive models typically suffer from stability issues. In order to mitigate this, we employ noise augmentation as well as an unrolling schedule, which are described in detail in the following sections. Unless stated otherwise, all models in this work, including baselines, are trained for $5\times 10^5$ iterations using the Adam optimizer with an exponential learning rate decay from $10^{-3}$ to $10^{-6}$.

\subsubsection{Unrolling Schedule}
\label{sec:methodology_us}
The original MeshGraphNets model
is trained by minimizing a mean squared error (MSE) loss computed between the predicted and ground-truth states at a single time step. However, the model is deployed autoregressively over many time steps during testing. This mismatch between training and inference motivates the use of unrolling during training, which improves the model's robustness to the accumulation of prediction errors.

\begin{figure}[pos=!ht]
    \centering
    \includegraphics[width=0.5\textwidth]{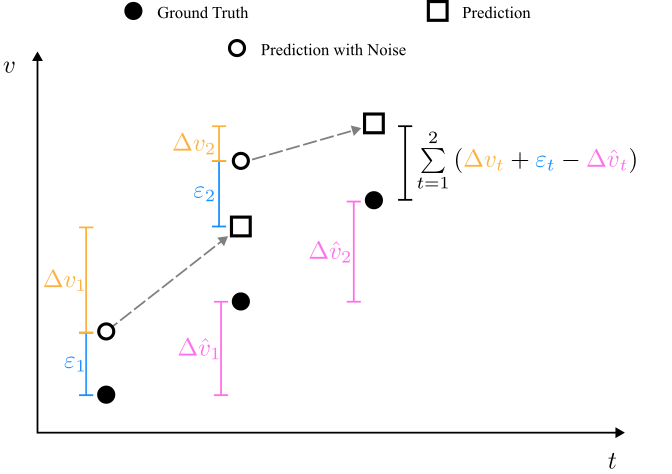}
    \caption{Schematic of unrolling procedure with two timesteps and assuming a node with only a one-dimensional feature. }
    \label{fig:unrolling_scheme}
\end{figure}

Figure \ref{fig:unrolling_scheme} exemplifies the unrolling procedure executed for two timesteps, one-dimensional node feature and scalar output field. At each timestep, noise $\varepsilon_t$ is added to the inputs (described in detail in Sec.~\ref{sec:noise}), and by the end of the rollout the error is calculated by
\begin{align}
    \bm{e}_m^i = \begin{bmatrix} \sum\limits_{t=1}^{m}  \Delta {{{v}}}_{t}^i + {\varepsilon}_{t}^i - \Delta \hat{v}_{t}^i   \\ s^i_m - \hat{s}^i_m \end{bmatrix} \mbox{ ,}
\end{align}
where $m$ is the number of timesteps employed, and $\Delta v_t^i$ and $\Delta \hat{v}_t^i$ are the model-estimated and true values of the scalar feature at timestep $t$. More precisely, we start the training  by unrolling the model for one timestep and linearly increase this window as the training advances. Increasing the horizon window linearly allows for the optimization landscape to slowly become more complex and alleviate training issues. Here, we use a hyperparameter $R$ to define the maximum unrolling length which will be reached at the end of training. Also, we use the error between predictions across all timesteps of the rollout. Therefore, we train the GNNs by minimizing the loss function

\begin{align}
& \mathcal{L}_k = \frac{1}{R_k} \sum_{m=1}^{R_k} \mathbb{E}\left[ \bm{e}_m^2 \right]
\\
& \text{where} \quad R_k = \min\left( \text{ceil}\left( \frac{k+1}{I} \cdot R \right),\; R \right)
\end{align}
where $k$ denotes the current training iteration (global step) and $I$ is the total number of training iterations. The quantity $R_k$ represents the number of unrolling steps used at iteration $k$, which increases linearly from $1$ to $R$ as training progresses.
The expectation $\mathbb{E}[\cdot]$ averages over the spatial nodes in the graph and output variables.
Note that the same amount of noise added to input variables is subtracted from the target labels in order to compensate for the modified input and a different amount of noise is used at each timestep.



\subsubsection{Noise Augmentation}
\label{sec:noise}

{
To prevent catastrophic drift during long autoregressive rollouts, we introduce a dynamic noise augmentation strategy applied to the input states. At each unrolling step, Gaussian noise is added to the physical properties stored on  nodes according to
}
\begin{align}
\bm{\varepsilon}^i = \beta \, h_{\varepsilon} \, \bm{\sigma}_y \odot \bm{\eta}^i, \quad \text{where} \quad \bm{\eta}^i \sim \text{Normal}(0,1), \;\; \beta \sim \text{Uniform}(0, 1), \label{eq:noise}
\end{align}
{
where $h_{\varepsilon}$ is a global scaling hyperparameter ($h_{\varepsilon} = 0.25$ in this work), $\bm{\sigma}_y$ is a vector containing the standard deviation of each variable across the ground-truth dataset, $\odot$ denotes element-wise multiplication, and $\bm{\eta}^i$ is drawn independently per node $i$.
}

{
The parameter $\beta$ is a single scalar sampled independently from a uniform distribution at each unrolled step and applied globally across all nodes. In practice, $\beta$ acts as global noise amplitude modulation without altering the spatial independence of $\bm{\eta}^i$.
By incorporating $\beta$ directly into the unrolled training loss, our methodology bridges two major stabilization concepts in the literature. When $\beta \to 0$, no explicit synthetic noise is injected ($\bm{\varepsilon}_t^i \to \mathbf{0}$). In this regime, the unrolled trajectory relies purely on model-generated errors, drawing inspiration from pushforward methods \cite{brandstetter2022message} where the network learns to correct its own accumulated drift. Conversely, when $\beta \to 1$, the input states experience maximum Gaussian perturbation, extending single-step noise injection strategies \cite{sanchez2020learning} into a multi-step unrolled setting. Dynamically sampling $\beta$ at each unrolled step exposes the model to a continuous spectrum of corruptions, forcing the GNN to build resilience against autoregressive drift.
}



\subsection{System Identification and Estimation}
\label{sec:estimation}

Besides conventional error metrics, we also evaluate different models by assessing their performance on system identification and estimation tasks. Here, we define a reference signal of a single flow property $\hat{\bm{q}} \in \mathbb{R}^{B \times T_s}$ taken at $B$ probe locations, where $T_s$ is the signal length, and an optimizable variable $\bm{\theta} \in \mathbb{R}^{V \times P+1 }$, which is scaled according to
\begin{align}
    \bm{\theta}_{\text{scal}} = 15 \bm{\sigma}_x \tanh{\bm{\theta}} \mbox{ ,}
\end{align}
where $\bm{\sigma}_x$ correspond to the standard deviations of the flow properties used as input to the model.
After scaling, $\bm{\theta}_\text{scal}$ is added to a random initial condition taken from the training dataset. More specifically, this initial condition is comprised of the physical features that evolve over time, such as $u_x$ and $u_y$ for the cylinder, and $u_x$, $u_y$ and $T$ for the cavity systems, as well as the non-dimensional parameters $\Rey$ and $\Ray$, respectively.
The model is then used autoregressively $T_s-1$ times and a loss is calculated as
\begin{align}
    \mathcal{L}_{\text{inv}} = \frac{1}{T_s}\sum_{t=1}^{T_s} \mathbb{E}\left[ (\bm{q}_t - \hat{\bm{q}}_t)^2 \right] \mbox{ ,} \label{eq:loss_inv}
\end{align}
where the expectation $\mathbb{E}[\cdot]$ averages over all probe nodes. It consists of computing the $L_2$ norm between the reference signal $\hat{\bm{q}}$ and the predicted snapshots sampled at all probe locations $\bm{q} \in \mathbb{R}^{B \times T_s}$.

\begin{figure}[pos=!ht]
    \centering
    \includegraphics[width=0.7\textwidth]{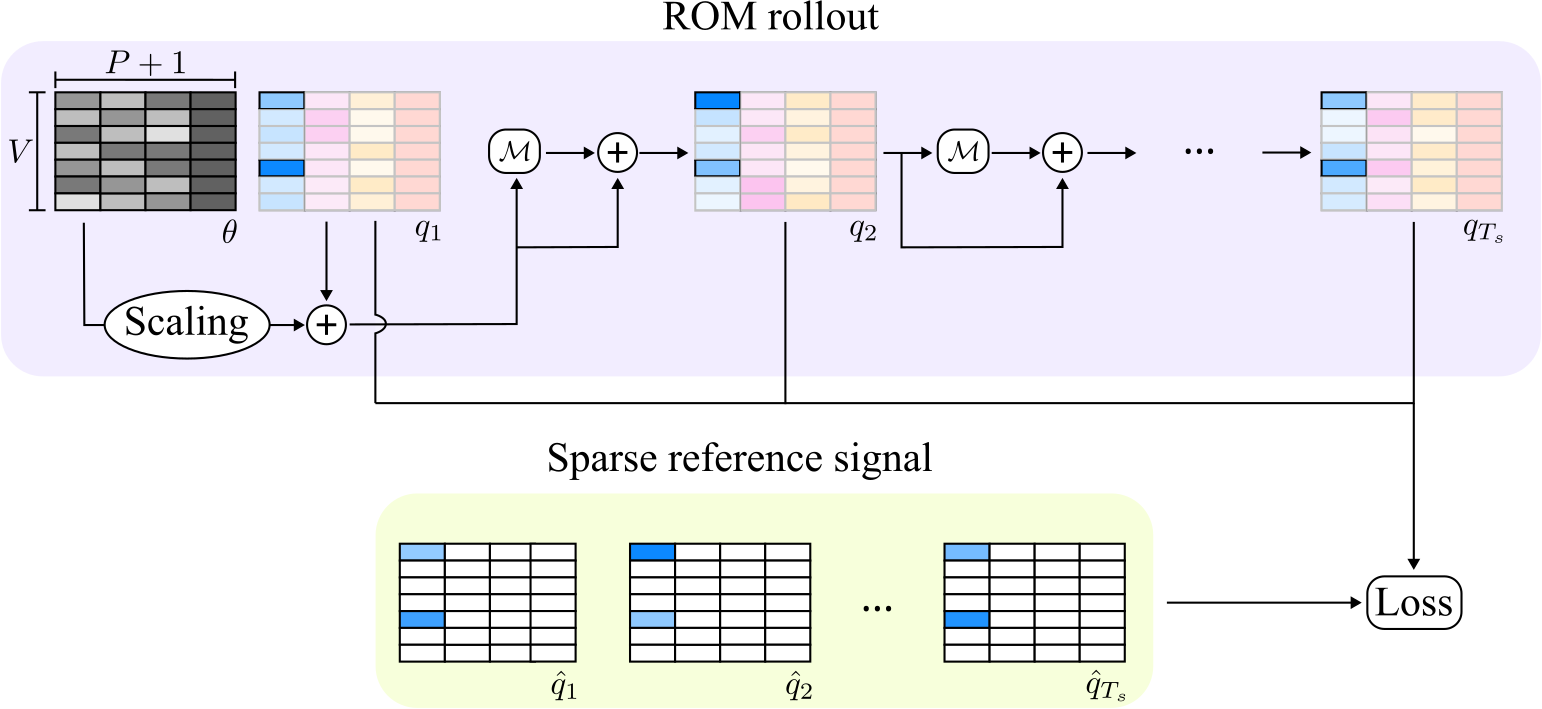}
    \caption{System estimation scheme. }
    \label{fig:inverse_scheme}
\end{figure}

Since the model is differentiable, it is possible to minimize the loss with respect to $\bm{\theta}$ while keeping the model weights frozen. This optimization yields an initial condition that reproduces the reference signal when marched forward in time. Consequently, other flow quantities are also reconstructed throughout the entire temporal evolution. Node attributes that represent either $\Rey$ or $\Ray$ are mapped to a single scalar across all nodes to facilitate the optimization.  A scheme illustrating this process can be seen in Fig.~\ref{fig:inverse_scheme}.

We found that a two-stage optimization procedure yields more accurate estimations. In the first stage, the optimization is carried out using  $T_s/2$ timesteps. The last predicted state is then used as the initial condition for a second optimization, which is performed over the remaining half of the reference signal. This strategy effectively initializes the latter optimization with a state closer to the true initial condition. For the cylinder case, reference signals are the horizontal velocity $u_x$, $T_s=40$, and optimizations are performed for $500$ iterations. For the cavity case, reference signals are temperature $T$, $T_s=60$, and optimizations are performed for $700$ iterations. Both systems use an exponential learning rate decay from $2\times 10^{-2}$ to $5 \times 10^{-4}$.


\section{Results}
\label{sec:results}
To evaluate the impact of the proposed techniques on the 2D testbeds, we train three models with different random seeds for each configuration and use their ensemble during testing. The surrogate simulation is marched in time from an unseen initial condition by feeding the same inputs to all models, averaging their outputs, and using the result to produce the next timestep.
For the 3D cylinder case, a single model per configuration is trained.

We compare models using the normalized mean square error $e_P$ described by
\begin{align}
    & e_P = \frac{ \mathbb{E} \left[ (\bm{y}_t^i - \hat{\bm{y}}_t^i)^2 \right]}{ \mathbb{E} \left[ (\hat{\bm{y}}_t^i)^2 \right] } \mbox{ ,}
\end{align}
where $\bm{y}_t^i = \begin{bmatrix} v_1^i & \cdots & v_P^i  & s^i \end{bmatrix}^T_t$ is a vector containing the predicted physical input features obtained by deploying the models autoregressively as well as the predicted scalar field. The respective ground truth values are represented by $\hat{\bm{y}}_t^i$ and only nodes at probe locations are considered.
This metric can then be averaged for all tests resulting in a value that reflects the overall model accuracy on probe points $\overline{e_P}$. Evaluating models using both metrics helps identifying the effects of the proposed techniques more clearly.


\subsection{Effect of Mesh Refinement}
\label{sec:refinement}

\begin{figure}[pos=!ht]
    \centering
    \begin{subfigure}[t]{\textwidth}
        \includegraphics[width=\textwidth]{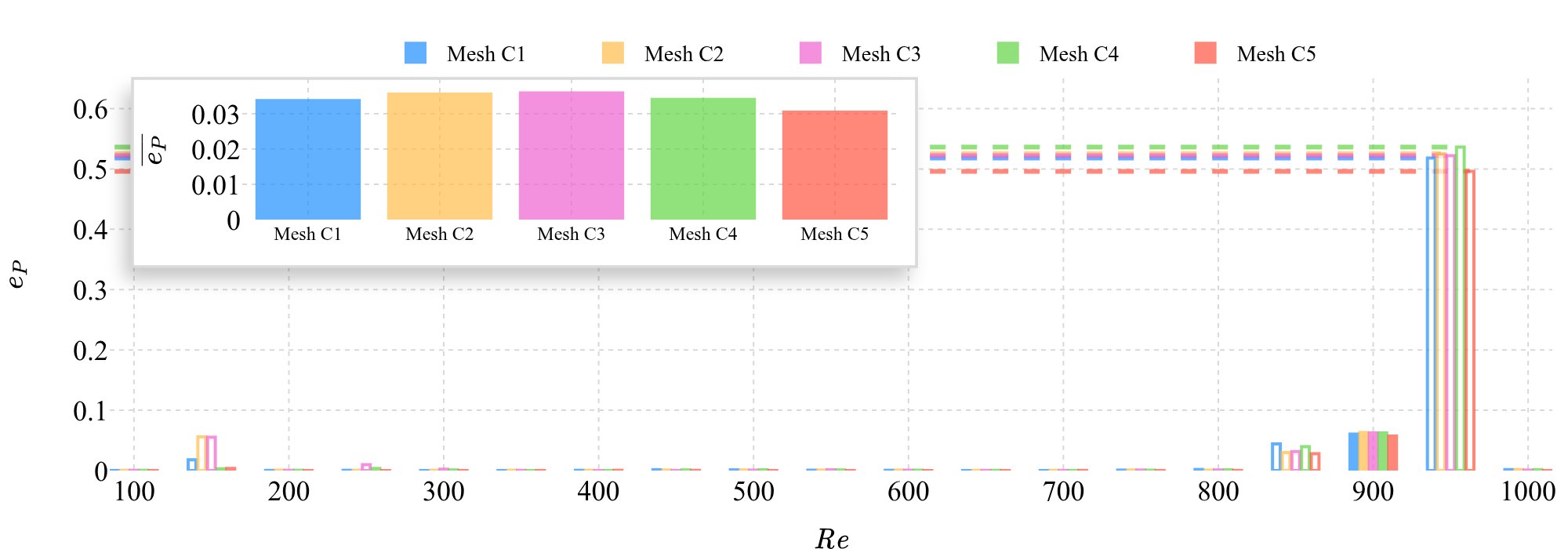}
    \end{subfigure}
    \begin{subfigure}[t]{\textwidth}
        \includegraphics[width=\textwidth]{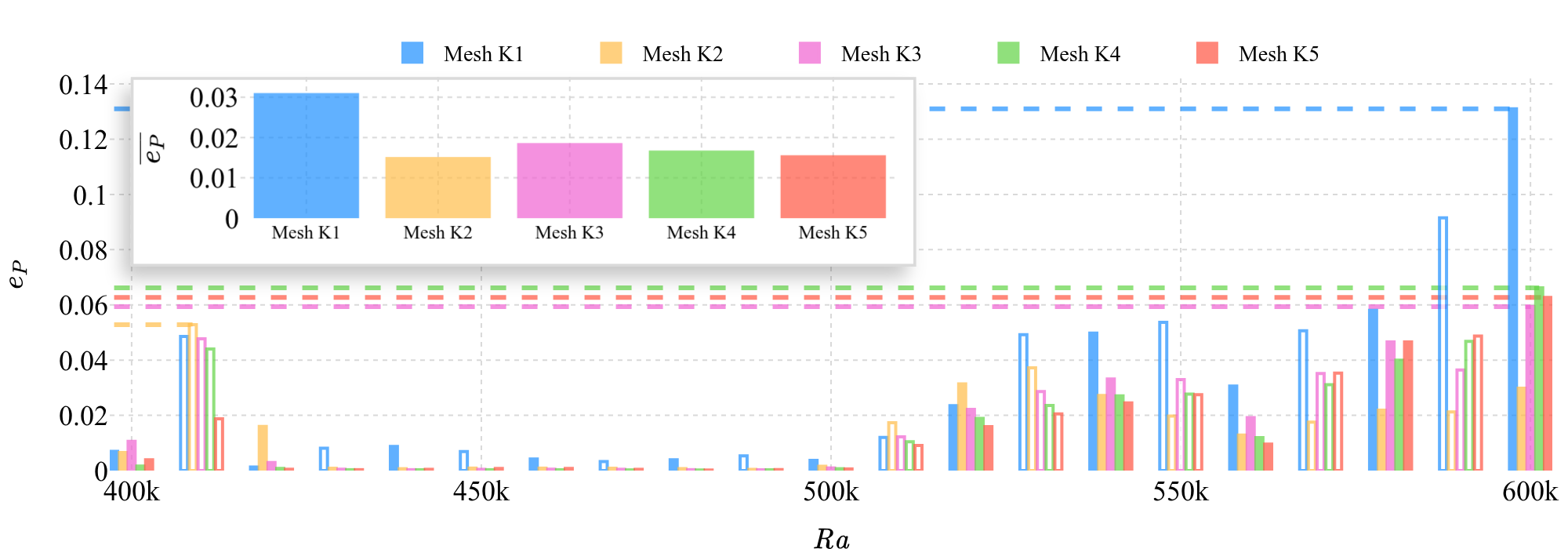}
    \end{subfigure}
    \caption{Cylinder (top) and cavity (bottom) surrogate errors by mesh size. Dashed lines represent maximum values of $e_P$. Filled and empty bars of $e_P$ represent tests with parameters present and not present in the training dataset, respectively.}
    \label{fig:mr}
\end{figure}

We begin our analysis by investigating the effect of mesh refinement on $e_P$ and $\overline{e_P}$ as shown in Fig.~\ref{fig:mr}.
{To ensure a fair comparison, we employ the same GNN architecture across all mesh resolutions. This is possible because local graph operations decouple the parameter count from the graph size. Consequently, the only variable across configurations is the mesh topology itself.}
For the cylinder case, all models exhibit very similar error trends, with the highest errors occurring at higher Reynolds numbers ($\Rey = 850, 900, 950$) with small differences between meshes. Error peaks also occur at $\Rey = 150$, where the von K\'{a}rm\'{a}n vortex street rapidly intensifies (as observed in Fig.~\ref{fig:cyl_dynamicsa}), which indicates that regions of the parameter space with changing dynamics are more difficult to learn. Interestingly, these results also demonstrate that increasing dynamical complexity does not necessarily imply higher errors. For example, the errors at $\Rey = 1000$ are considerably lower than the ones at $\Rey=900$ for all models. This behavior contrasts with classical numerical methods, where dynamical complexity typically increases the difficulty of achieving accurate solutions.

A similar trend is observed for the cavity flow, with the highest errors concentrated at the second half of the parameter space, where dynamics are considerably more chaotic, but also at $\Ray=4.1\times 10^5$. In this region of the parameter space, property fluctuations become out of phase between the hot and cold wall, representing a distinct shift in the dynamical behavior despite the flow retaining its periodic nature. For this system, however, the mesh refinement level has a greater impact on overall model accuracy. There is a decrease of more than half $\overline{e_P}$ for models trained on Mesh K2 compared to the models trained on Mesh K1. Increasing the level of refinement from this point does not further decrease $\overline{e_P}$ constistently across all cases; conversely,  there is an increase in the maximum values of $e_P$ observed at higher values of \Ray.
This result shows that models trained on coarser meshes can actually perform better than those trained on finer ones, which contradicts common intuition from classical numerical methods. This is particularly significant given that, in absolute terms, all models already achieve very low prediction errors across the vast majority of the parameter space.

These results indicate that after a minimum level of refinement, model accuracy stalls or slightly degrades. For a simpler system such as the confined cylinder flow, the coarsest mesh employed is already sufficient. For the cavity flow, which presents richer dynamics, one level of refinement is sufficient to achieve an accuracy plateau.

It could be expected that training with very coarse graphs would lead to oversmoothing \cite{rusch2023survey}.
A key component to avoid this problem is the use of skip connections on  nodes and edges processors.
When training models without them, their outputs collapse to the same values for all nodes.
%
Another important point is that each iteration of training uses a whole graph for updating the model weights. Therefore, training with denser graphs is analogous to using larger batch sizes, which could hinder the training process at some point.
Recent works attempt to circumvent these issues by employing multiscale graphs \cite{fortunato2022multiscalemeshgraphnets, BARWEY2023112537} or masking \cite{ICLR2025_9e443458}. Here, we adopt a coarse mesh, as our primary objective is to obtain predictions at specific probe locations and to compute integrated quantities commonly used in engineering practice.

It is also worth pointing out that training with coarse meshes allow for smaller datasets,
which can facilitate sharing and storage. Other attractive considerations are faster training and less memory usage. While not shown, it was also observed that training on the coarsest and finest meshes resulted in models more sensitive to the amount of noise used during training, requiring finer tuning of this hyperparameter.

The receptive field is another factor for model accuracy that can be affected by mesh size.
Its extent is determined by mesh refinement and the number of message-passing blocks $L$ utilized by a model. For example, for the same $L$, a finer mesh in practice has a smaller receptive field, making information from farther away unavailable. If the receptive field is too small, potentially useful information is not accessible for predictions, decreasing performance. To understand this effect, models are trained with different values of $L$ while keeping the total number of trainable parameters constant.

\begin{figure}[pos=!ht]
    \centering
    \begin{subfigure}[t]{0.49\textwidth}
        \includegraphics[width=\textwidth]{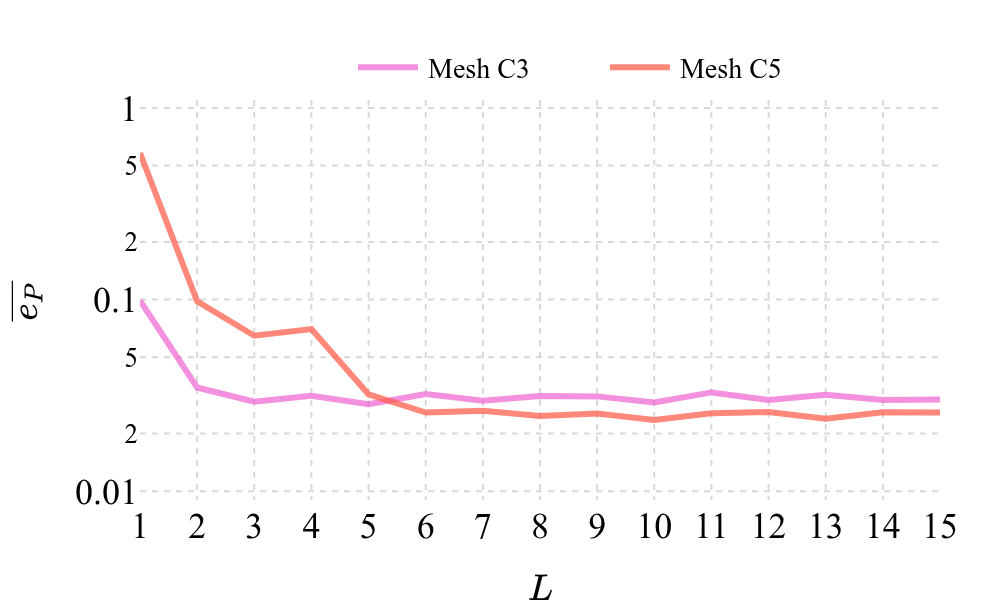}
    \end{subfigure}
    \begin{subfigure}[t]{0.49\textwidth}
        \includegraphics[width=\textwidth]{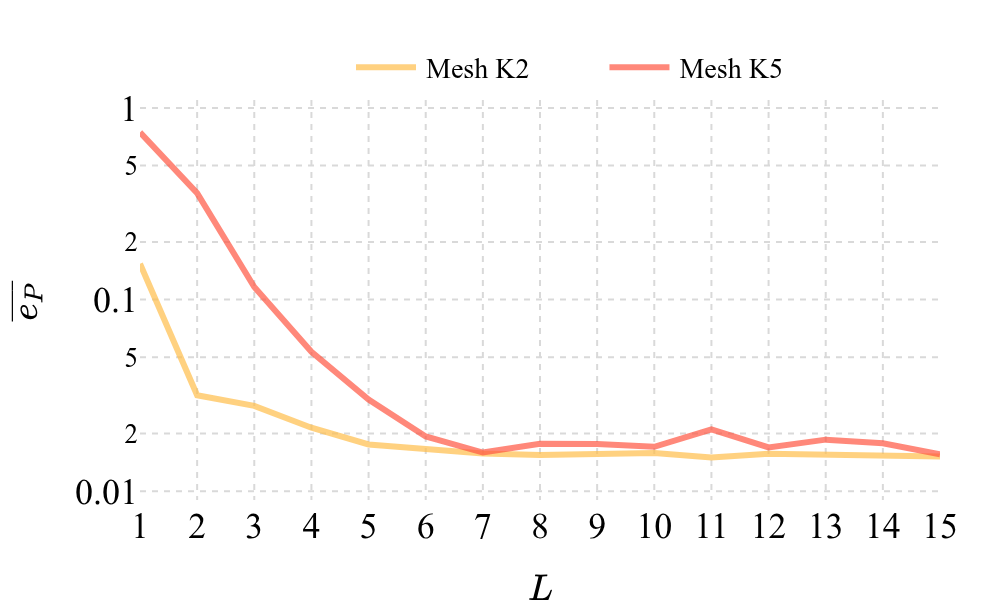}
    \end{subfigure}
    \caption{Effect of the number of message-passing blocks on $\overline{e_P}$ for the cylinder (left) and cavity (right) flows. All models have approximately the same number of trainable parameters.}
    \label{fig:nsteps_ep}
\end{figure}

Results show a convergence behavior of $\overline{e_P}$ after a certain receptive field size is achieved, as shown in Fig.~\ref{fig:nsteps_ep}. The coarse meshes C3 and K2 converge at $L \approx 3$, while finer ones (C5 and K5) reach an accuracy plateau at $L\approx 6$. This behavior indicates that there is likely a minimum receptive field size required to achieve adequate accuracy.

\begin{figure}[pos=!ht]
    \centering
    \begin{subfigure}[t]{1\textwidth}
        \centering
        \includegraphics[width=\textwidth]{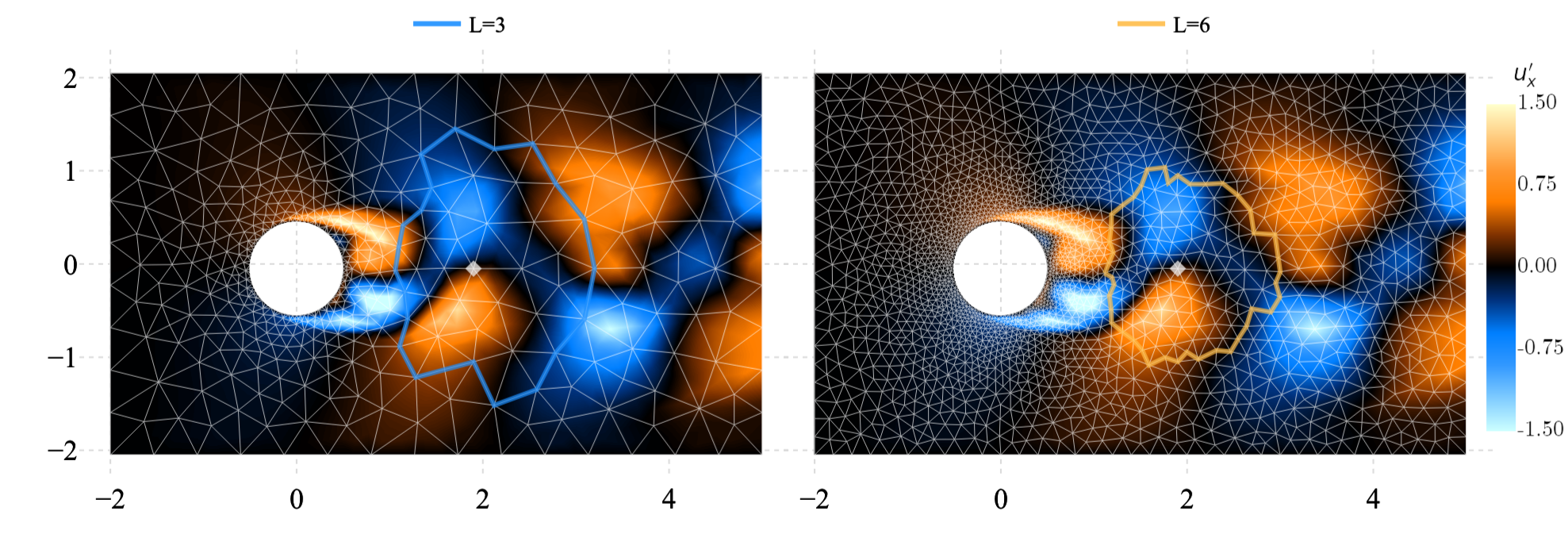}
    \end{subfigure}
    \caption{Receptive field sizes for Mesh C3 (left) and Mesh C5 (right). Contours represent $u_x$ fluctuations at $\Rey=1000$.}
    \label{fig:nsteps_cyl}
\end{figure}

\begin{figure}[pos=!ht]
    \centering
    \begin{subfigure}[t]{0.42\textwidth}
        \centering
        \includegraphics[width=\textwidth]{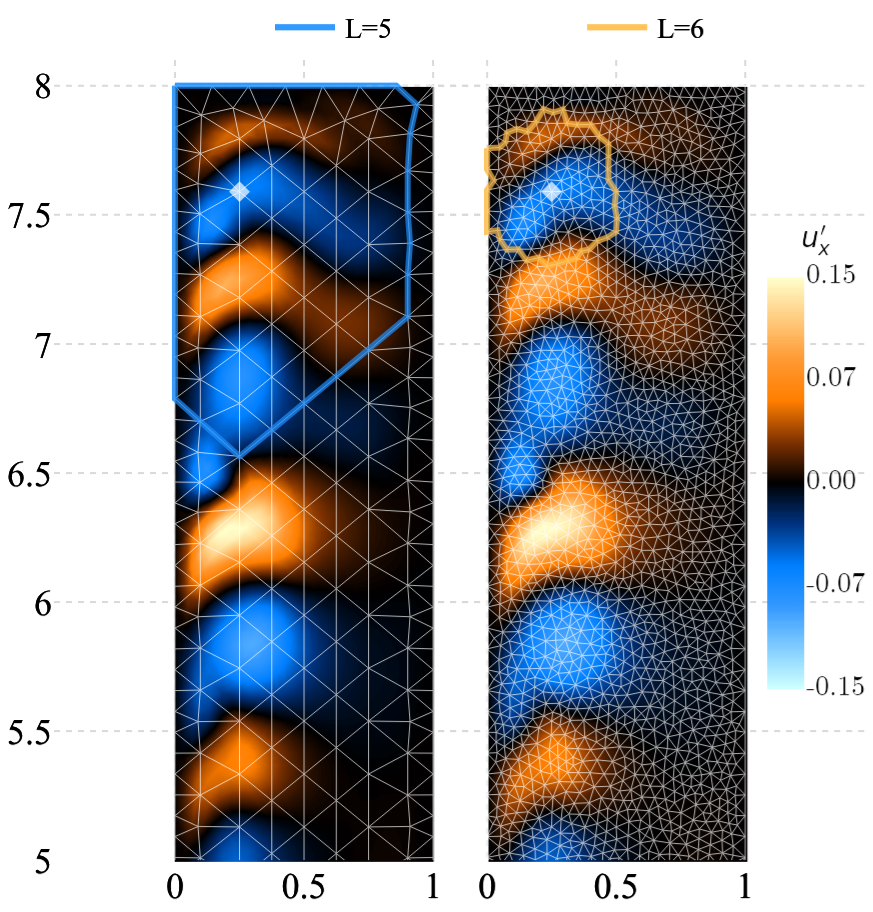}
    \end{subfigure}
    \caption{Receptive field sizes for Mesh K2 (left) and Mesh K5 (right). Contours represent $u_x$ fluctuations at $\Ray=600k$.}
    \label{fig:nsteps_cav}
\end{figure}

To understand how the receptive field size might relate to flow physics, different receptive fields are shown in Figs.~\ref{fig:nsteps_cyl} and \ref{fig:nsteps_cav} for the cylinder and cavity cases, respectively. Their centers are located at probe coordinates with the highest error with overlaid contours of $u_x$ fluctuations. Convergence appears to be reached when one to two wavelengths of relevant coherent structures are covered. This suggests a potential heuristic for selecting $L$ given a mesh resolution, though further validation on other flows is required.

\subsection{Effect of Non-Uniform Training Dataset}

\begin{figure}[pos=!ht]
    \centering
    \begin{subfigure}[t]{\textwidth}
        \includegraphics[width=\textwidth]{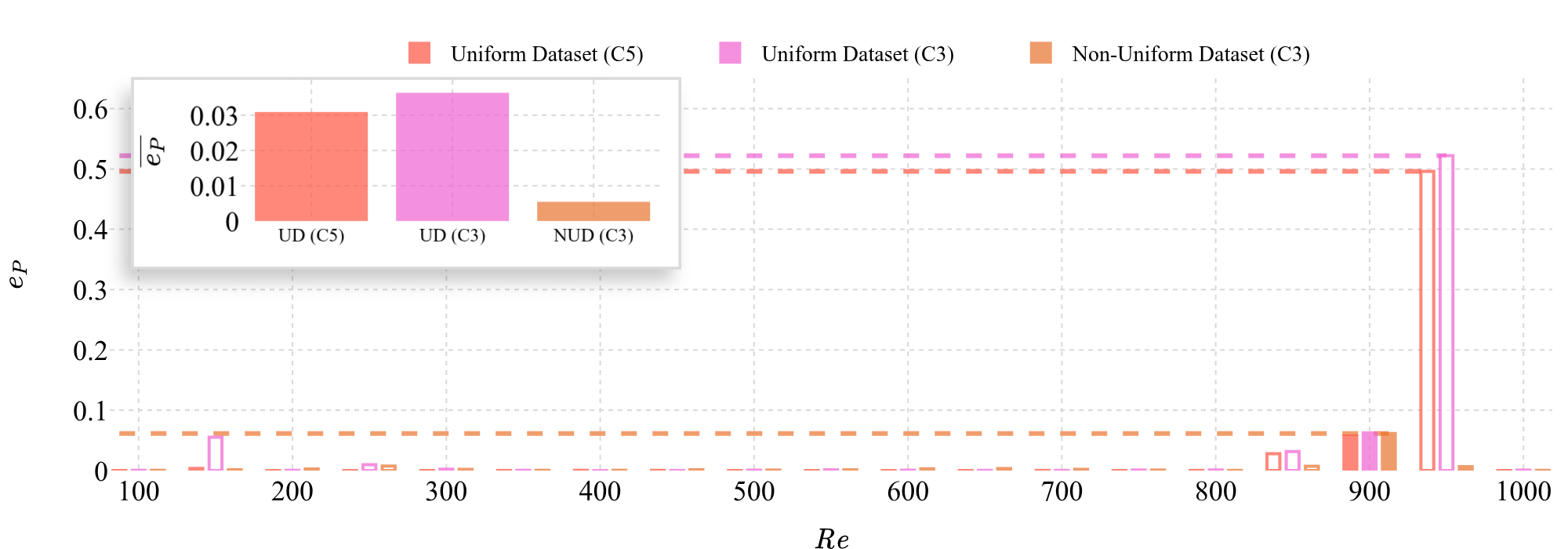}
    \end{subfigure}
    \begin{subfigure}[t]{\textwidth}
        \includegraphics[width=\textwidth]{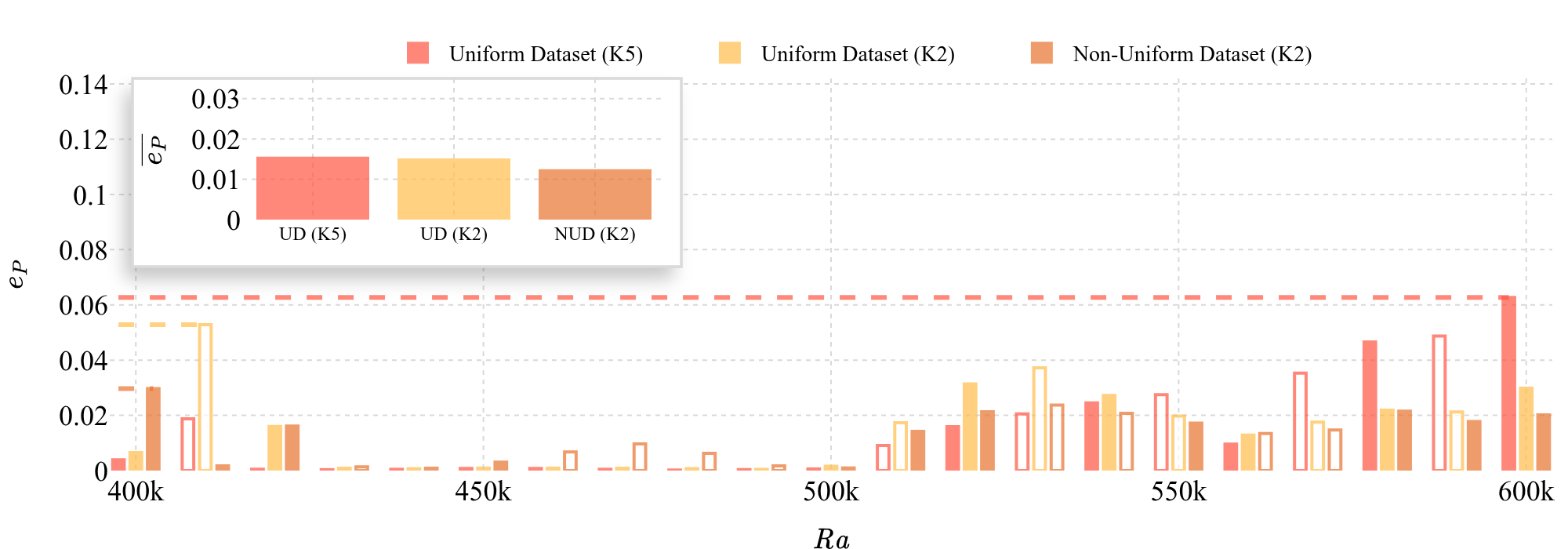}
    \end{subfigure}
    \caption{Cylinder (top) and cavity (bottom) surrogate model errors for different training datasets. Dashed lines represent maximum values of $e_P$. Filled and empty bars of $e_P$ represent tests with parameters present and not present in the training dataset, respectively.}
    \label{fig:nud}
\end{figure}

It is clear from the results presented in Section \ref{sec:refinement} that errors are not homogeneously distributed across the parameter space. This motivates the use of a training dataset non-uniformly sampled in parameter space, as presented in Sec.~\ref{sec:methodology_nud}. The effect of employing this technique is shown in Fig.~\ref{fig:nud}, with Mesh C3 and Mesh K2 being used. All datasets contain the same number of training instances and vary only in the sampling distribution over the parameter space.

For the cylinder case, the maximum errors are significantly reduced, particularly at $\Rey=950$, where $e_P$ decreases by a factor of approximately two orders of magnitude. This suggests that concentrating samples in regions where the dynamics vary most rapidly can improve accuracy. For the cavity flow, the maximum $e_P$ also decreases, with particularly pronounced reductions at $\Ray=4.1 \times 10^5$. In regions of the parameter space where samples are sparse, errors tend to increase, as expected. However, these are regions where model accuracy is already very high, so a slight degradation is greatly compensated by the error reduction in regions of poorer performance. Note that the time required to run the algorithm presented in Sec.~\ref{sec:methodology_nud} is just a minimal fraction of that needed for a single training run. Thus, this methodology enhances model accuracy without significant computational cost.


\subsection{Effect of Unrolling Schedule}
\begin{figure}[pos=!ht]
    \centering
    \begin{subfigure}[t]{\textwidth}
        \includegraphics[width=\textwidth]{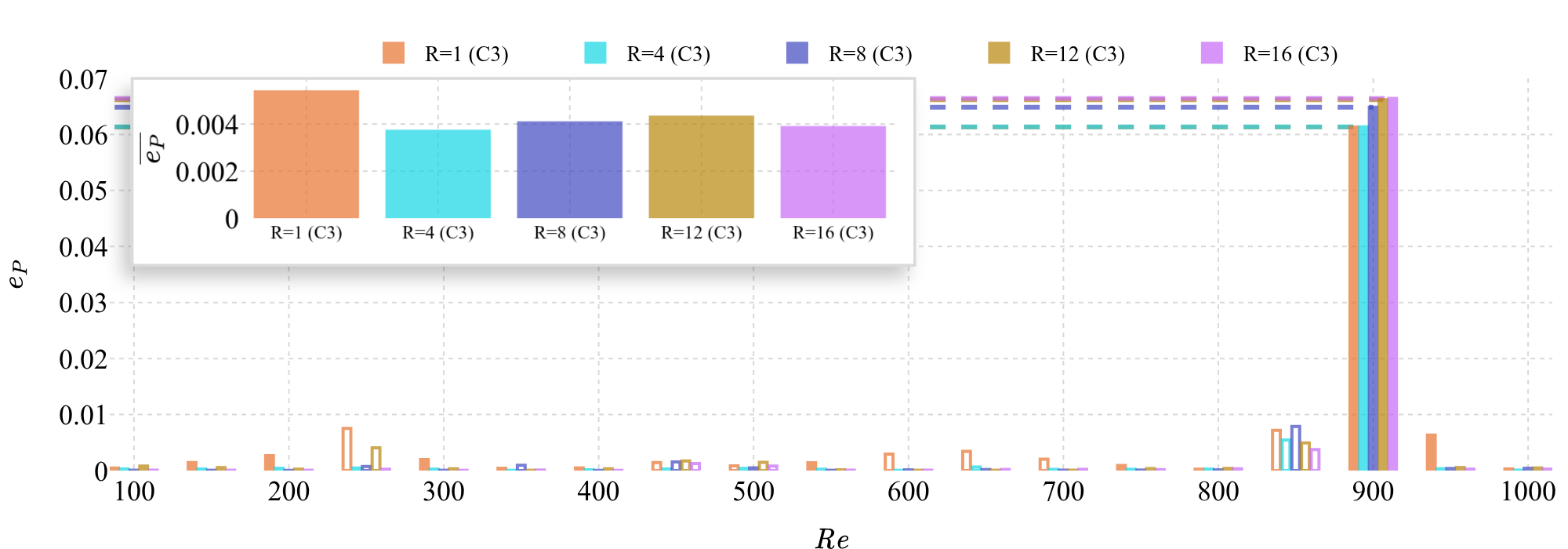}
    \end{subfigure}
    \begin{subfigure}[t]{\textwidth}
        \includegraphics[width=\textwidth]{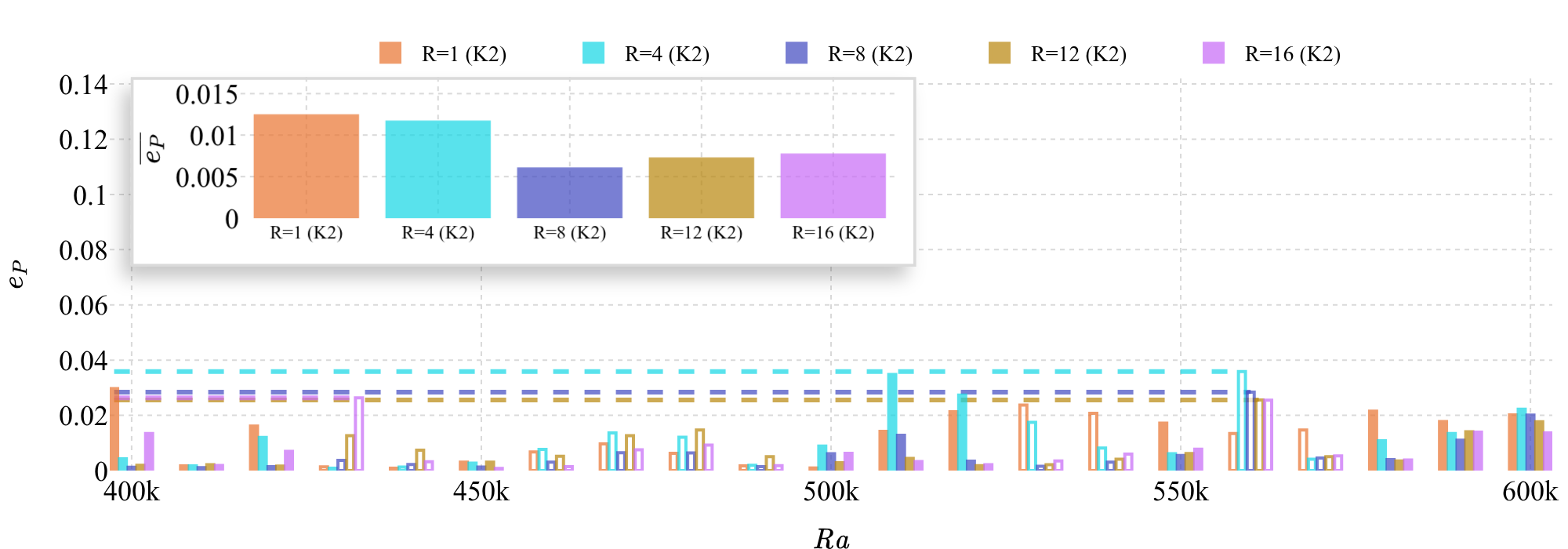}
    \end{subfigure}
    \caption{Errors for the cylinder (top) and cavity (bottom) cases for different unrolling horizons. Bars represent maximum and minimum values of $e_{P}$. Filled bars correspond to test parameters seen during training but with different initial conditions.}
    \label{fig:rl}
\end{figure}

In order to mitigate error accumulation due to autoregressive deployment, we use the unrolling during training scheduling presented in Sec.~\ref{sec:methodology_us} together with non-uniform training datasets. The effect of using different horizons $R$ is presented in Fig.~\ref{fig:rl}. Overall, accuracy improves until reaching a plateau, which happens at $R=4$ and $R=8$ for the cylinder and cavity flows, respectively. The first exhibits a $31\%$ reduction in $\overline{e_P}$, while a $50\%$ reduction is observed for the second. When compared to models trained on the finest meshes with uniform sampling, $\overline{e_P}$ is reduced by $60\%$ and $88\%$ for the cylinder and cavity cases, respectively. Thus, model accuracy is greatly enhanced by using the proposed unrolling schedule.
Interestingly, accuracy on the test with $\Rey=900$ for the cylinder flow remains unaffected by any of the techniques presented in this work. In this region of the parameter space, the flow begins to exhibit a broader frequency spectrum, which may not be as easily learned as other regions. This highlights the importance of analyzing model accuracy not only by a single global integrated metric. Improvements in the cavity flow are more pronounced for $\Ray > 5\times 10^5$, where errors are generally higher and less prone to model-to-model variability, although there are individual test cases, such as at $\Ray=4.3\times 10^5$, where $e_P$ increased.

\begin{figure}[pos=!ht]
    \centering
    \includegraphics[width=0.5\textwidth]{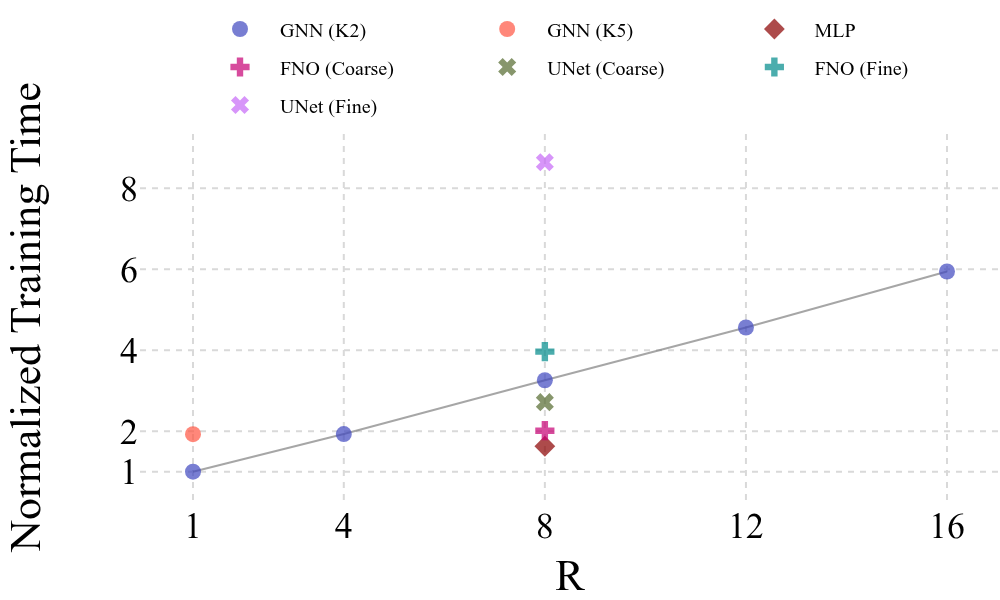}
    \caption{Training time as a function of unrolling horizon. Values are normalized by the time for $R=1 \text{(K2)}$.}
    \label{fig:rl_time}
\end{figure}

Training with longer unrolling horizons has two main trade-offs: increased memory usage and training time. The first is largely mitigated by employing coarser meshes. The second, however, remains relevant because the speedups gained from using coarser meshes plateau beyond a certain refinement level, and other bottlenecks begin to dominate training time. Figure~\ref{fig:rl_time} shows time linear dependency of $R$. Training times for baseline methodologies (further discussed in Sec.~\ref{sec:baseline}) are also presented. Although training time increases with $R$, training with $R=4$ on Mesh K2 takes approximately the same time as training with $R=1$ on Mesh K5. Thus, using a coarser mesh enables more accurate models within the same computational cost.

\subsection{Qualitative Comparison of Time-Series Predictions}

\begin{figure}[pos=!ht]
    \centering
    \includegraphics[width=\textwidth]{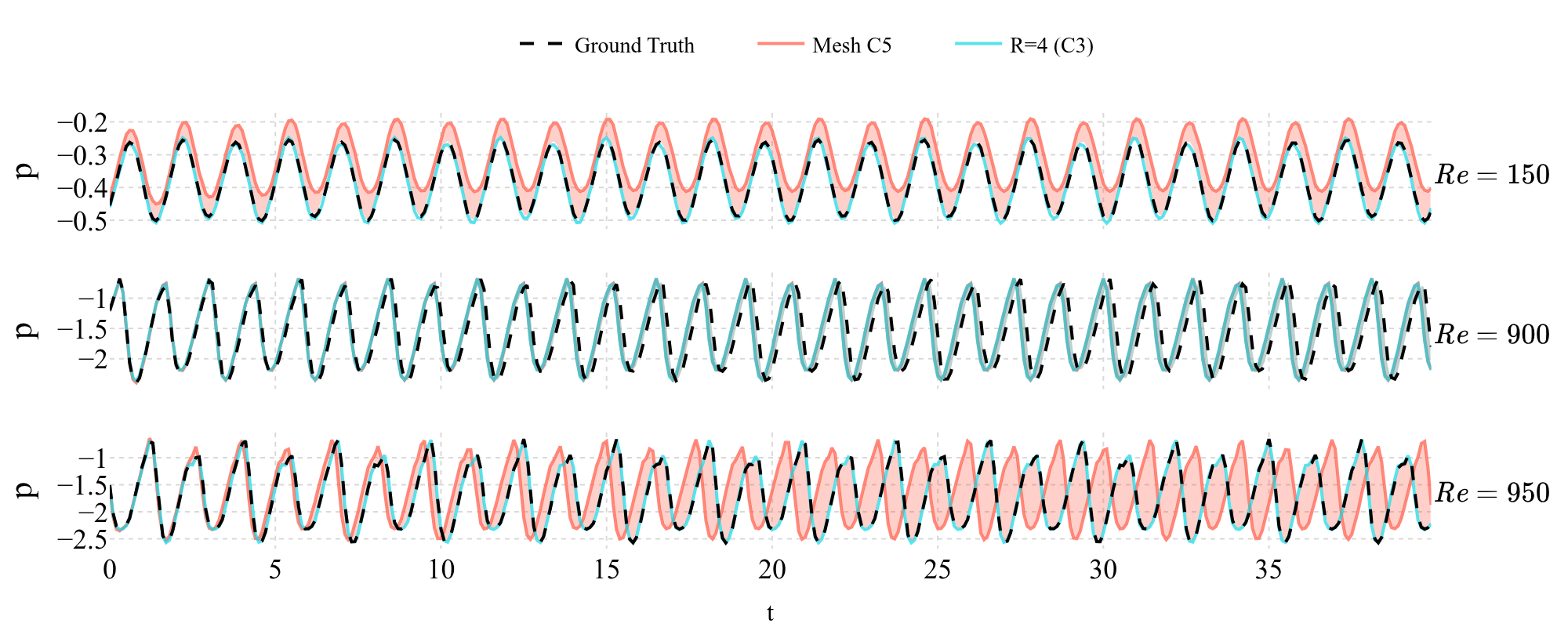}
    \caption{Probe signals at $(x,y) = (0.6, -0.05)$ for cylinder flow at various Reynolds numbers.
    Phase shifts are significantly reduced by the proposed approach incorporating both mesh-coarsening and unrolling.}
    \label{fig:probes_cyl}
\end{figure}

To better visualize the error reduction reported in previous sections, we compare time-series signals at locations exhibiting the highest errors in both parameter space and spatial domains. Figure~\ref{fig:probes_cyl} shows these comparisons for the cylinder flow case, where errors primarily manifest as phase shifts. They are greatly diminished when using the proposed techniques: a coarse mesh (Mesh C3) with $R=4$ and non-uniformly sampled training data. As previously noted, $\Rey=900$ poses an interesting case where prediction accuracy remains unchanged across all models. For this case, results corresponding to meshes C5 (R1) and C3 (R4) are not distinguishable in the plot. On the other hand, unrolling considerably improves the prediction for $\Rey=950$, matching the phase and amplitude with the ground truth solution.

\begin{figure}[pos=!ht]
    \centering
    \includegraphics[width=\textwidth]{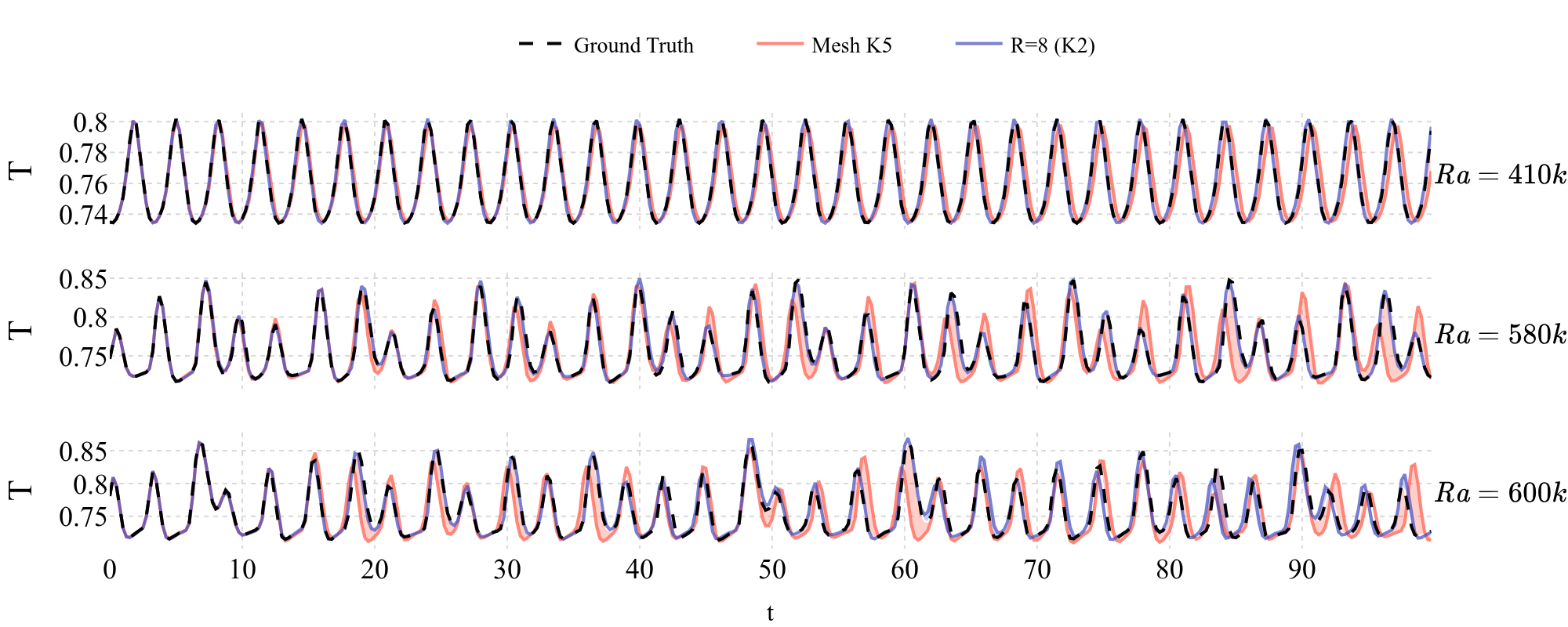}
    \caption{Probe signals at $(x,y) = (0.25, 7.59)$ for cavity flow at various Rayleigh numbers. 
    The proposed method maintains accuracy even as the flow transitions from periodic to intermittent behavior.}
    \label{fig:probes_cav}
\end{figure}

Figure~\ref{fig:probes_cav} presents corresponding signals for the cavity flow. At lower Rayleigh numbers where the flow is fully periodic, errors are again dominated by phase shifts. However, as $\Ray$ increases, the flow loses periodicity and exhibits intermittent patterns, making the predictions considerably more challenging. The proposed techniques successfully mitigate errors across the entire parameter space, including these difficult transitional regimes.

\begin{figure}[pos=!ht]
    \centering
    \includegraphics[width=\textwidth]{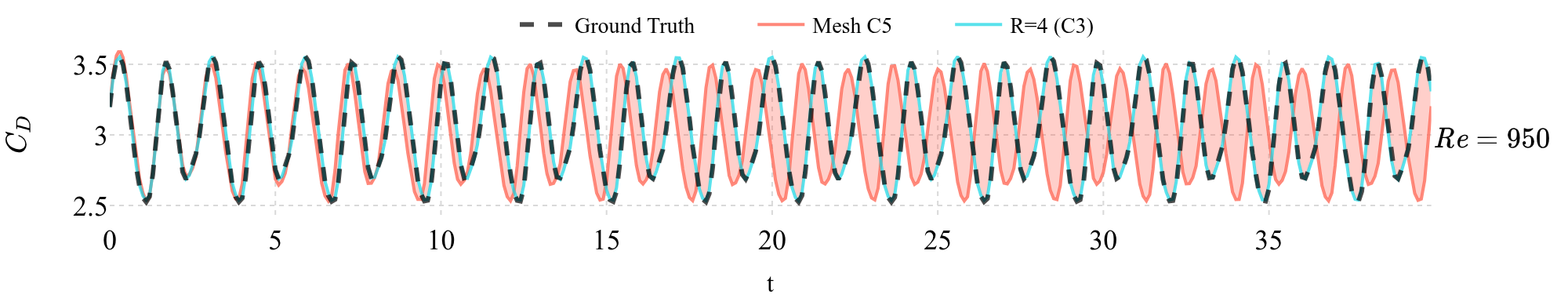}
    \caption{Drag coefficient $(C_D)$ comparison for cylinder flow.}
    \label{fig:scalars_cyl}
\end{figure}

\begin{figure}[pos=!ht]
    \centering
    \includegraphics[width=\textwidth]{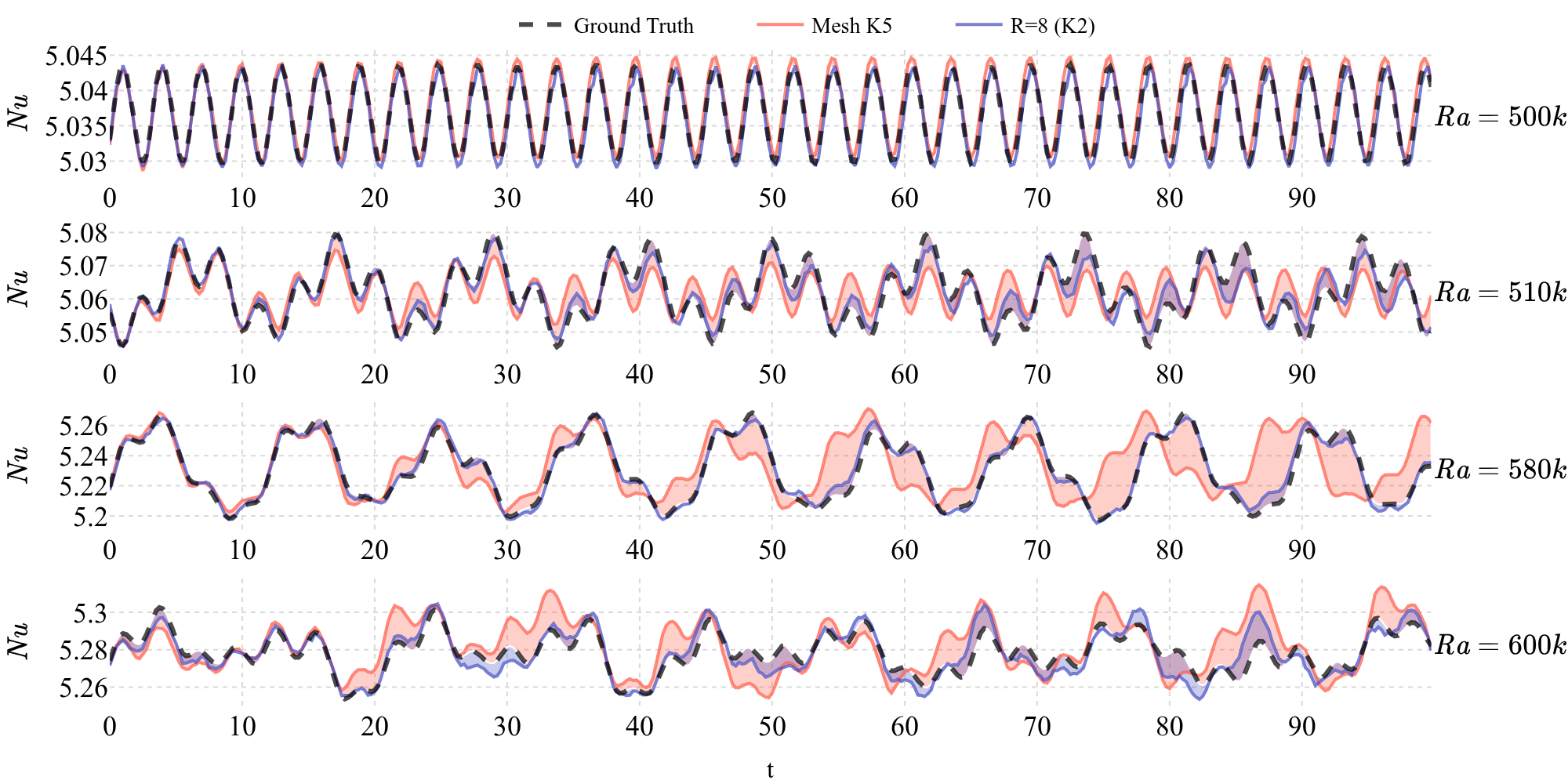}
    \caption{Nusselt number ($\textit{Nu}$) comparison for cavity flow. The proposed method K2 (R8) captures the transition from periodic to intermittent dynamics more accurately than the baseline setup K5 (R1).}
    \label{fig:scalars_cav}
\end{figure}

Figures~\ref{fig:scalars_cyl} and~\ref{fig:scalars_cav} present the predictions of drag coefficient $C_D$ and Nusselt number $\textit{Nu}$ for the cylinder and cavity flows, respectively. The most notable improvements for the cylinder case appear at $\Rey=950$, where phase shift errors are more pronounced. At other Reynolds numbers, the improvements are more modest, as the baseline model already provides fairly accurate predictions.

The cavity flow results displayed in Figure~\ref{fig:scalars_cav} show more substantial differences owing to their richer dynamics. At $\Ray=5.1\times 10^5$, the baseline model (mesh K5) rapidly converges to a single frequency limit cycle likely because the flow at $\Ray=5 \times 10^5$ exhibits a clear dominant frequency. In contrast, the proposed method ($R=8$, non-uniform sampling with unrolling schedule) accurately captures the dynamical shift. At higher Rayleigh numbers, where the dynamics is composed of several frequencies including intermittency, models trained with the proposed enhancements exhibit better phase and amplitude agreement with the reference solution compared to the baseline model.

%
\subsection{Baselines Comparison} \label{sec:baseline}

To evaluate the efficacy of the proposed coarse-mesh GNN framework, we compare its performance against a diverse set of baseline surrogates representing both linear modal techniques and standard deep learning operators. A parametric Dynamic Mode Decomposition (DMD) model \cite{HUHN2023111852} serves as a linear benchmark operating directly on probe and integration nodes. To evaluate a non-spatial baseline operating on sparse observables, Multilayer Perceptrons are trained on the same sparse state dataset similar to \cite{deda2023backpropagation, deda2024neural, Déda_Ramos_Wolf_Dawson_2026}. Furthermore, U-Nets \cite{ronneberger2015} and Fourier Neural Operators (FNOs) \cite{li2021fourier} are evaluated on both coarse grids (matching the spatial resolution of Mesh C3/K2) and fine grids (matching Mesh C5/K5). Because these structured architectures require regular grids that do not conform to unstructured probe locations, predicted fields are post-processed using cubic spline interpolation for comparison. The FNO is restricted to the cavity flow, as standard FNO architectures require rectangular tensor grids and cannot natively accommodate domains with internal obstacle cutouts. Physics-informed and projection-based surrogates are omitted, as the extreme sparsity of the graphs prevents the accurate discrete computation of spatial differential operators. All nonlinear baselines are trained under identical conditions using an unrolling horizon of $R=8$, the proposed non-uniform dataset sampling, and matching noise injection levels. Architecture details of the baselines are provided in Appendix~\ref{app:baselines}.

\begin{figure}[pos=!ht]
    \centering
    \begin{subfigure}[t]{\textwidth}
        \includegraphics[width=\textwidth]{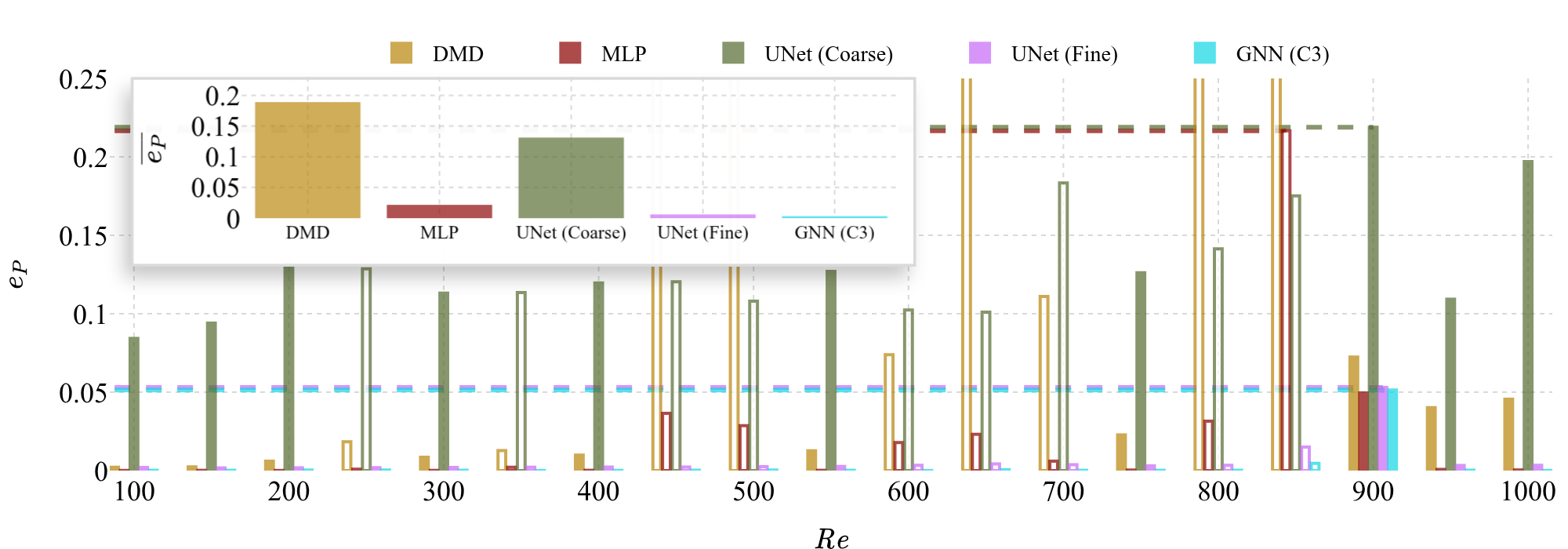}
    \end{subfigure}
    \begin{subfigure}[t]{\textwidth}
        \includegraphics[width=\textwidth]{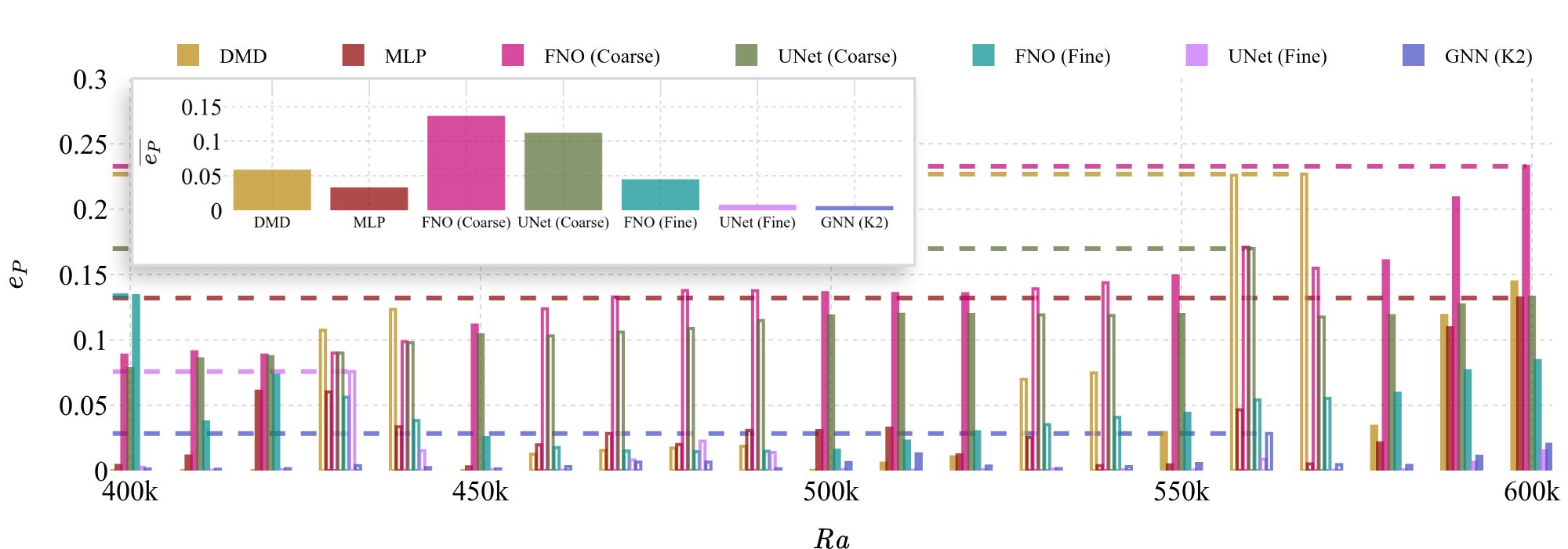}
    \end{subfigure}
    \caption{Errors for the cylinder (top) and cavity (bottom) cases for different types of models. Bars represent maximum and minimum values of $e_{P}$. Filled bars correspond to test parameters seen during training but with different initial conditions.}
    \label{fig:baselines}
\end{figure}

Results obtained for all models are summarized in Fig.~\ref{fig:baselines}. Parametric DMD achieves low prediction errors in periodic near-linear flow regimes at lower $\Rey$ and $\Ray$ numbers. However, its linear formulation degrades rapidly when generalizing to nonlinear regimes and parameter values absent from the training set. The sparse MLP achieves a lower overall mean error $\overline{e_P}$ than DMD, but exhibits high error spikes on unseen test parameters, highlighting its limited capacity for parametric generalization.

For structured operators (U-Net and FNO), restricting execution to coarse spatial grids severely degrades performance, rendering less accurate models than linear DMD in several test cases. This degradation stems from both resolution limitations and spatial interpolation errors at non-conforming probe sites. On fine grids, U-Net accuracy improves significantly, almost matching the low error levels of the coarse-mesh GNN. However, fine-mesh U-Nets require predicting many more spatial degrees of freedom, incurring substantial computational memory overhead and training latency. While fine-mesh FNO performance improves over its coarse counterpart, it remains prone to autoregressive error accumulation during long time-horizon rollouts.

Overall, the coarse-mesh GNN achieves lower prediction error than all sparse and coarse-mesh baselines while matching the predictive accuracy of fine-mesh operators at a fraction of their training cost. Training times for all baselines except DMD are reported in Fig.~\ref{fig:rl_time} (DMD is omitted because it requires negligible execution time). Coarse-mesh baselines are more computationally efficient than GNNs on comparably sized graphs because dense operations, convolutions, and FFTs are better optimized on GPUs than graph operations. However, the only baseline with comparable accuracy, U-Net (Fine), requires more than twice as long to train. These results demonstrate the advantages of localized graph message passing for modeling sparse flow observables directly on unstructured spatial topologies and for generalizing the dynamics across the full training parameter distribution.

%
\subsection{System Identification and Estimation Assessment}

\begin{figure}[pos=!ht]
    \centering
    \begin{subfigure}[t]{0.4\textwidth}
        \includegraphics[width=\textwidth]{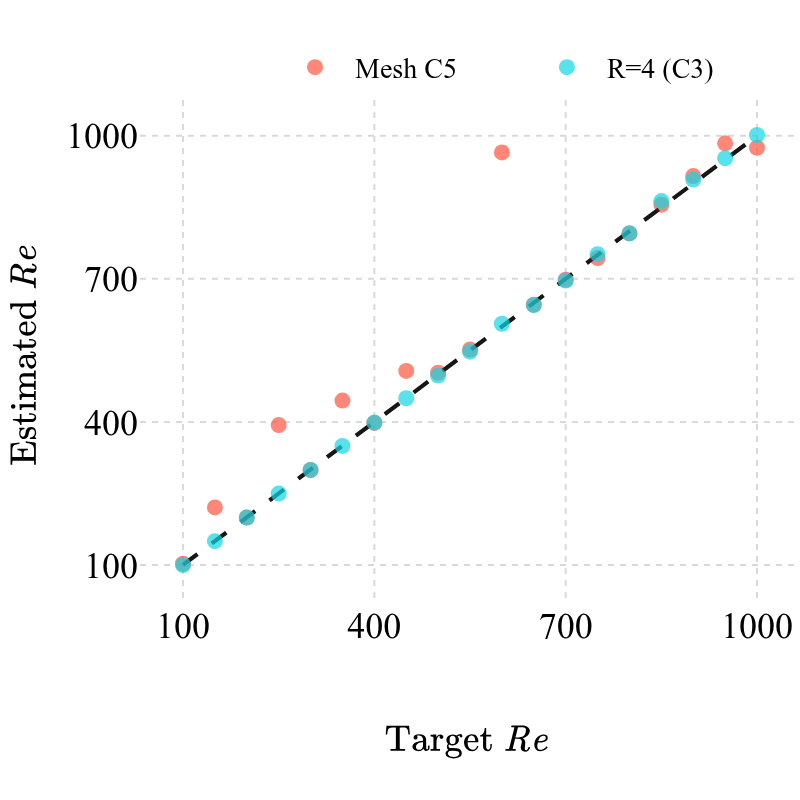}
    \end{subfigure}
    \hspace{8mm}
    \begin{subfigure}[t]{0.4\textwidth}
        \includegraphics[width=\textwidth]{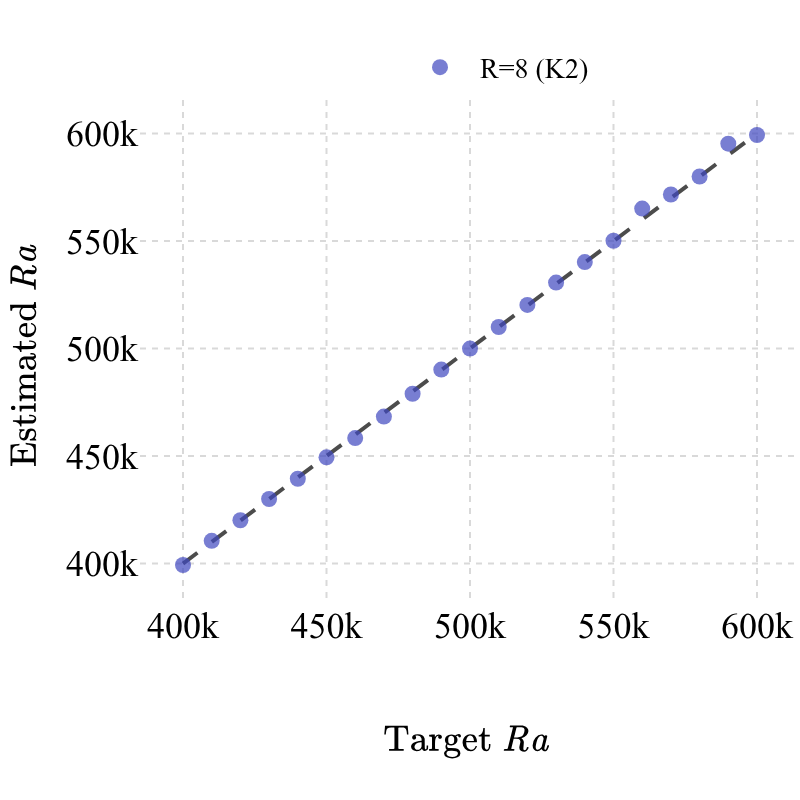}
    \end{subfigure}
    \caption{Estimation of $\Rey$ (left) and $\Ray$ (right) from various sparse reference signals across the parameter space.}
    \label{fig:scalar_estimation}
\end{figure}

Models are also evaluated according to their capability of performing system identification and estimation tasks, where an optimization is performed to find an initial condition that when marched in time matches reference signals at probe locations. Only a single model is used and not an ensemble, as the previous sections, since there were no significant differences between models trained with different seeds.
Figure~\ref{fig:scalar_estimation} shows estimations of $\Rey$ and $\Ray$ for signals across the entire parameter space. For the cylinder case, models trained on Mesh C5 tend to exhibit a more unstable optimization process, likely due to the higher amount of degrees of freedom, resulting in considerably less accurate estimates compared to $R=4$ (C3), which exhibits a maximum estimation error of $1.4\%$ of the parameter space length. When considering the cavity flow, performing the estimation utilizing Mesh K5 is not possible due to memory constraints. Nevertheless, estimations using $R=8$ (K2) exhibit low errors, with the highest being only $2.5\%$ of the parameter space range.

\begin{figure}[pos=!ht]
    \centering
    \begin{subfigure}[t]{0.49\textwidth}
        \includegraphics[width=\textwidth]{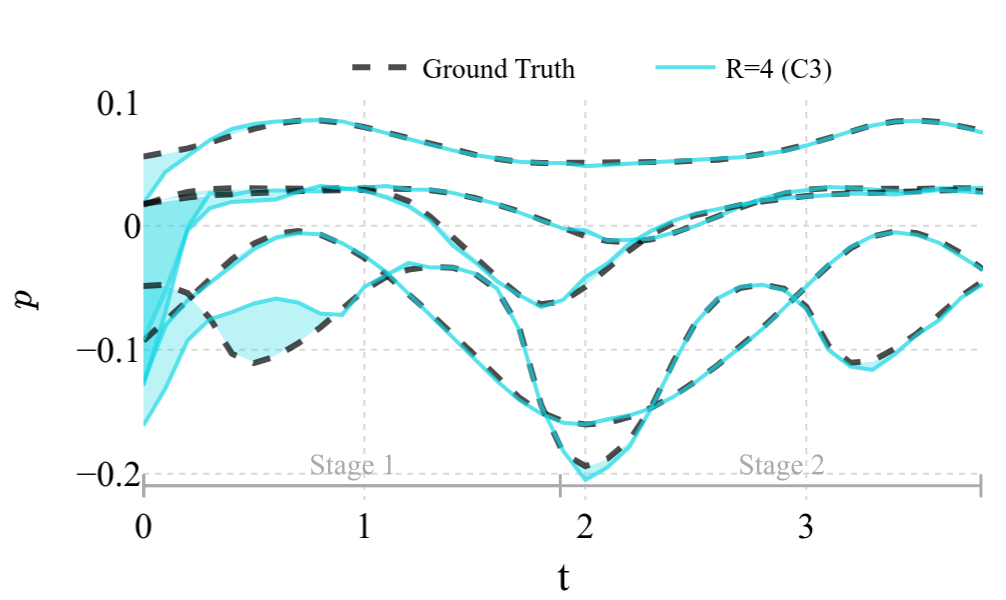}
    \end{subfigure}
    \begin{subfigure}[t]{0.49\textwidth}
        \includegraphics[width=\textwidth]{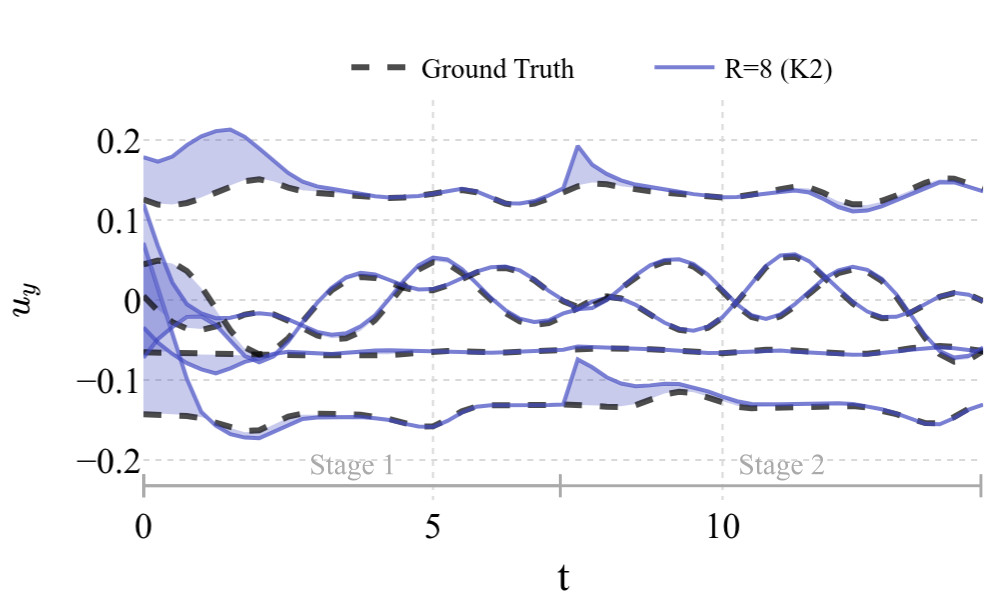}
    \end{subfigure}
    \caption{Estimation of $p$ and $u_y$ for the cylinder (left) and cavity (right) cases, respectively. Reference signals are acquired at $\Rey=900$ and $\Ray=6\times 10^5$, respectively.}
    \label{fig:probes_estimation}
\end{figure}

This optimization process aims to find an initial condition that, when evolved over time with the model, matches the sparse reference signal. Since this is done purely in a data-driven fashion, there is no guarantee that the obtained initial condition is physically plausible. This behavior can be seen in Fig.~\ref{fig:probes_estimation}, where estimated signals from various probes  are shown. Although the obtained initial condition is not physically accurate, after some timesteps predictions quickly match the ground truth values, especially after stage 2, effectively yielding accurate estimations.

These results demonstrate that models employing the proposed techniques are capable of accurately recovering system parameters from sparse observations across the entire parameter space. In contrast, models trained on conventional finer meshes either suffer from inaccurate estimations due to a more complex optimization landscape or demand a prohibitive amount of memory.


\subsection{Extension to 3D Parameterized Flow}

Having established and validated the core principles of the proposed techniques on the 2D benchmark flows, we assess if coarse-mesh GNNs can also perform well in a 3D setting. Comparisons are performed between GNNs trained on three meshes with increasing levels of refinement. For this case, tests are performed using a single model per configuration and not an ensemble of models as in the previous sections due to the higher computational cost of training.

\begin{figure}[pos=!ht]
    \centering
    \begin{subfigure}[t]{\textwidth}
        \includegraphics[width=\textwidth]{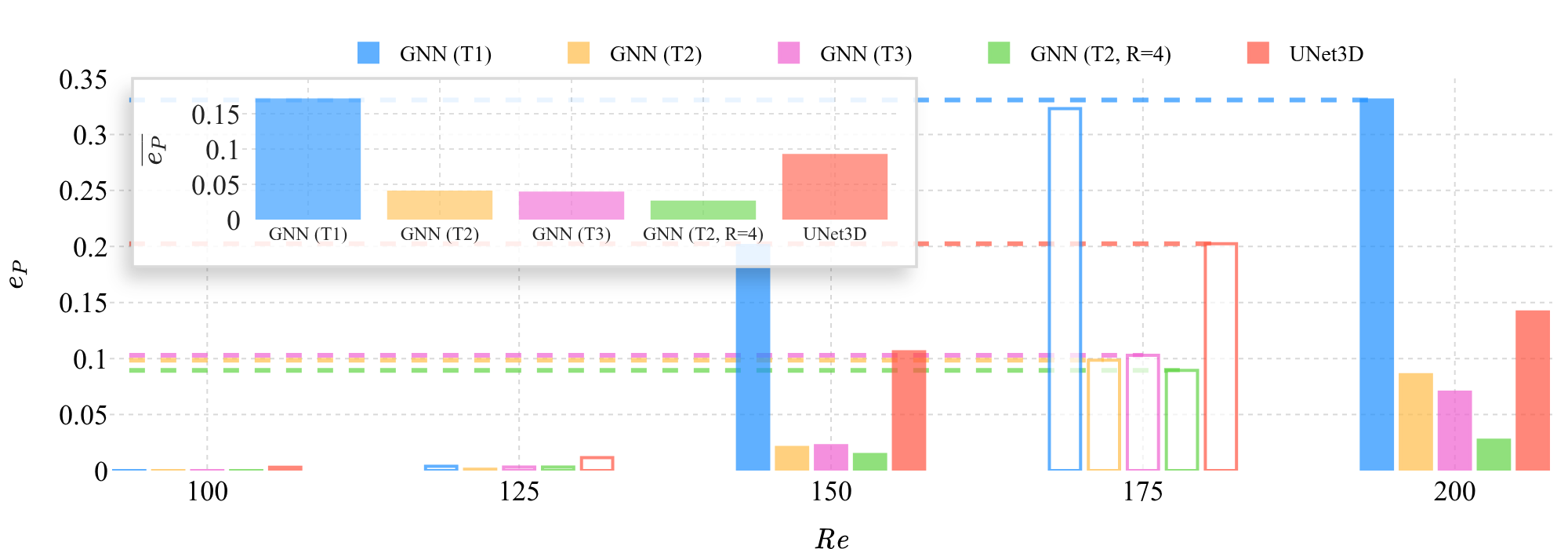}
    \end{subfigure}
    \caption{Errors for the 3D cylinder case for different mesh refinement levels and  models. Bars represent maximum and minimum values of $e_{P}$. Filled bars correspond to test parameters seen during training but with different initial conditions.}
    \label{fig:error_cyl3D}
\end{figure}

A summary of the errors is shown in Fig.~\ref{fig:error_cyl3D}. All models perform well on the 2D regime ($\Rey<150$) and deviations become stronger when 3D effects are present.  When it comes to using finer meshes for training the GNNs, the same behavior as in the 2D cases is observed where a plateau in $\overline{e_P}$ is reached after a minimum level of refinement. Thus, instead of spending more computational resources training models on finer meshes, a more efficient strategy is to use unrolling during training. For this case, the GNN trained on mesh T2 with unrolling shows a reduction of $35\%$ in $\overline{e_P}$.

We consider previously employed baselines for this case, such as parametric DMD and an MLP, which are trained directly on data from probe locations, and a U-Net that utilizes 3D convolutions trained on a structured grid with approximately the same amount of points as mesh T3. All nonlinear baselines are trained with $R=4$. Only results from the U-Net are shown in Fig.~\ref{fig:error_cyl3D} since neither the parametric DMD and MLP were able to produce minimally accurate rollouts, with the exception of DMD when applied with $\Rey=100$. The U-Net produces an $\overline{e_P}$ three times higher than the GNN trained on the mesh T2 (half the amount of points) and $R=4$. It is likely that, as previously observed for the 2D testbeds, a considerably finer grid is necessary for the U-Net to achieve similar levels of accuracy.

\begin{figure}[pos=!ht]
    \centering
    \begin{subfigure}[t]{\textwidth}
        \includegraphics[width=\textwidth]{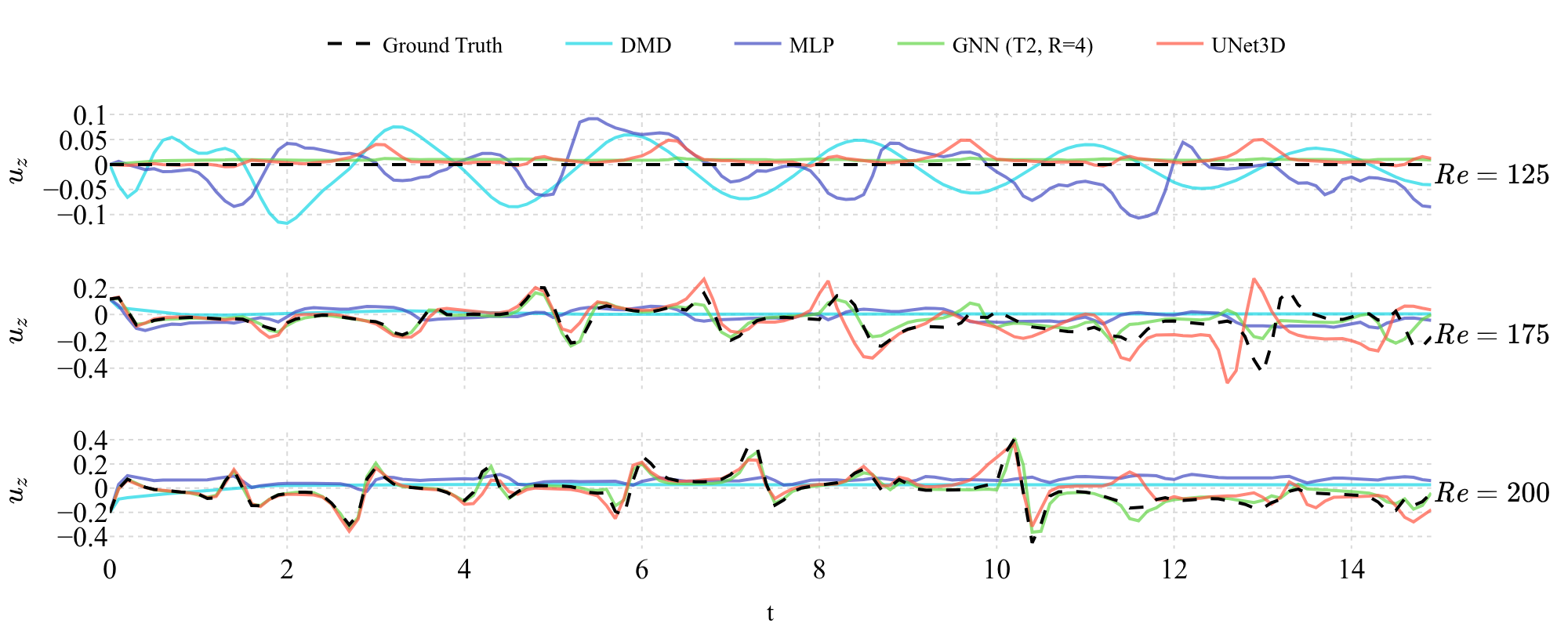}
    \end{subfigure}
    \caption{$u_z$ signals at $(x,y) = (3.2, -0.05)$ for 3D cylinder flow at various $\Rey$ numbers. 
    }
    \label{fig:signals_cyl3D}
\end{figure}

\begin{figure}[pos=!ht]
    \centering
    \begin{subfigure}[t]{\textwidth}
        \includegraphics[width=\textwidth]{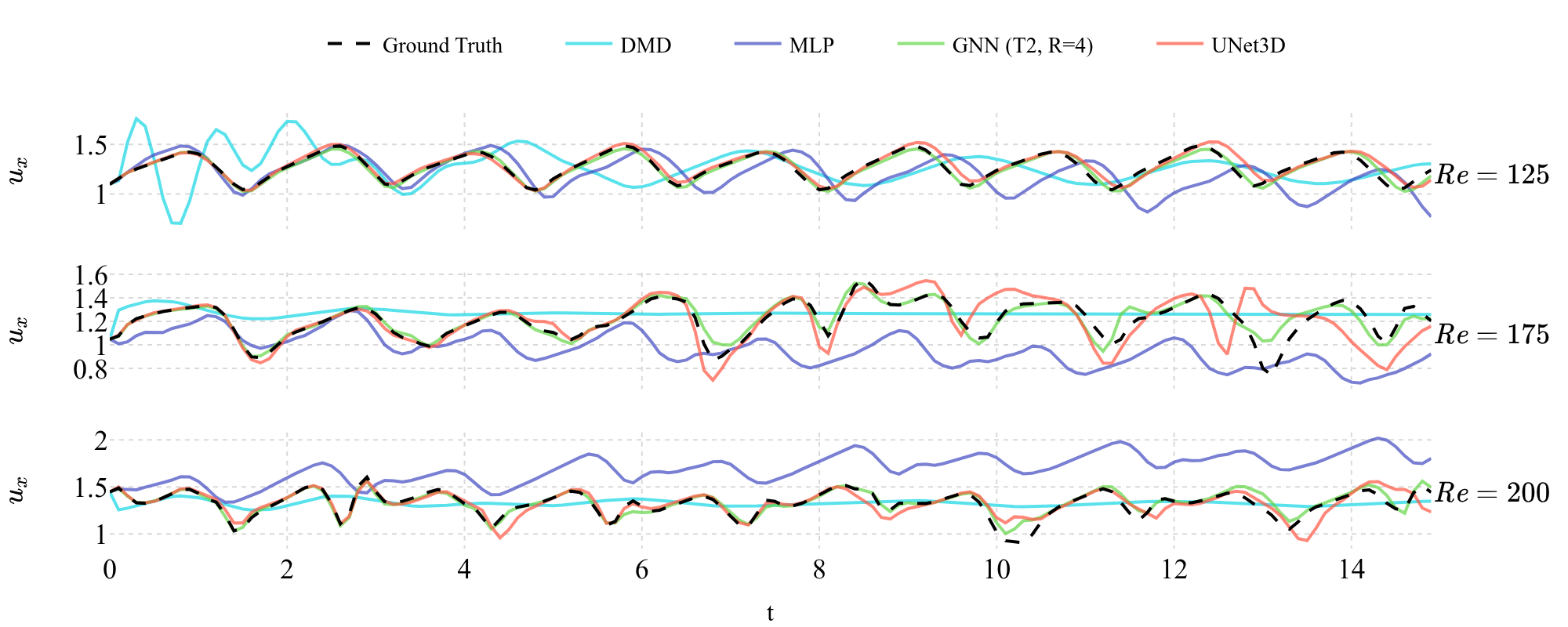}
    \end{subfigure}
    \caption{$u_x$ signals at $(x,y) = (3.2, -0.05)$ for 3D cylinder flow at various $\Rey$ numbers. 
    }
    \label{fig:signals_cyl3D2}
\end{figure}

The predicted signals of a single probe are displayed in Figs.~\ref{fig:signals_cyl3D} and \ref{fig:signals_cyl3D2}. The GNN (T2, R=4) model predictions are considerably more accurate and track the ground-truth solution closely much longer than the baselines. Even though training dataset samples are available only for $\Rey=(100, 150, 200)$, the GNN predictions at $\Rey=175$ only start to deviate after 12 convective time units. Deviations when considering the DMD and MLP based models rapidly increase after a few timesteps and, while the U-Net is the best baseline, it is considerably less accurate than the coarse-mesh GNN model.

\section{Conclusions}
\label{sec:conclusions}

This work investigates the use of graph neural networks for reduced order modeling of parameterized fluid flows by employing mesh coarsening, adaptive parameter-space sampling, and unrolling during model training. Through systematic evaluation of 2D and 3D unsteady flows depicting periodic, intermittent and chaotic dynamics, we demonstrate that these combined approaches yield significant accuracy improvements while offering computational advantages over 
linear methods and established machine-learning baselines.
These benefits are particularly relevant for reduced-order models that target only a subset of the domain degrees of freedom, such as probe measurements or integrated quantities. Collectively, these findings advocate for a shift from the default preference for high-resolution training data toward physics-informed choices that balance accuracy, efficiency, and practicality.

Our mesh refinement study reveals that finer meshes do not necessarily produce more accurate neural surrogate models, a counterintuitive finding that challenges established norms in computational physics. For the 2D cylinder flow, there was no substantial difference in accuracy between meshes, while for the cavity and 3D cylinder flow, only a single refinement level was required to reach an accuracy plateau. We observed that model accuracy is highly dependent on the receptive field size, with accuracy converging after approximately one to two wavelengths of dominant flow structures are covered. Thus, it is possible to obtain same levels of accuracy by utilizing a coarse mesh with a smaller number of messaging passing blocks.

The non-uniform training dataset strategy, guided by how the frequency spectrum changes across the parameter space, proved particularly effective in reducing maximum errors without significant computational overhead. For the cylinder flow, concentrating samples in regions of rapidly changing dynamics reduced maximum test error by an order of magnitude, while for the cavity flow, substantial improvements were observed at various regions of the parameter space. Importantly, the algorithm required to design these non-uniform datasets incurs only a fraction of the cost of a single training run, making this approach highly practical.

The unrolling schedule during training greatly benefits autoregressive deployment, reducing overall mean errors considerably, especially in regions of high dynamical complexity. The linear relationship between training time and unrolling horizon presents a practical trade-off which can be compensated by combining coarser meshes with moderate unrolling. Therefore, more accurate models can be trained with the same computational budget.

We also show how the proposed techniques can be beneficial for system identification and estimation tasks. By reducing the dimensionality of the learned dynamics, we not only circumvent the prohibitive memory and computational costs associated with high-resolution discretizations, but also facilitate a more stable and tractable optimization landscape for estimation. Notably, even when the estimated initial conditions lack immediate physical interpretability, the models rapidly converge to accurate predictions within a few time steps, confirming that the learned reduced dynamics successfully capture the essential system behavior required for downstream inference tasks.

The primary limitation of the proposed framework is its restriction to the parametric bounds defined by the training dataset, as predictive accuracy degrades rapidly when extrapolating beyond the trained Reynolds and Rayleigh number regimes. Furthermore, the network architecture assumes fixed, a priori known probe topologies. Thus, altering sensor placement or count requires updating the graph structure and possibly retraining the model. Additionally, the proposed heuristic linking network receptive field size to characteristic spatial flow wavelengths remains empirical and requires further validation over a broader range of physical systems.


Future work should focus on extending the proposed approach to setups with actuation and parametric surfaces, providing a natural pathway toward practical ROMs for control design and shape optimization. We anticipate that these techniques will be particularly valuable in engineering applications where computational budgets are constrained and accurate long-term predictions are required.


\section*{Acknowledgements}

The authors acknowledge the financial support received from Fundação de Amparo à Pesquisa do Estado de São Paulo, FAPESP, under Grants No. 2013/08293-7, 2021/06448-0, 2024/21444-9 and 2025/10531-0, and from Conselho Nacional de Desenvolvimento Científico e Tecnológico, CNPq, under Grants No. 444329/2024-2 and 304320/2024-2. STMD acknowledges support for US DOE grant DE-SC0025597.

\appendix
\section{Convergence Analysis and Mesh Refinement}
\label{app:convergence}
\begin{figure}[pos=ht!]
    \centering
    \begin{subfigure}[t]{0.49\textwidth}
        \centering
        \includegraphics[width=\textwidth]{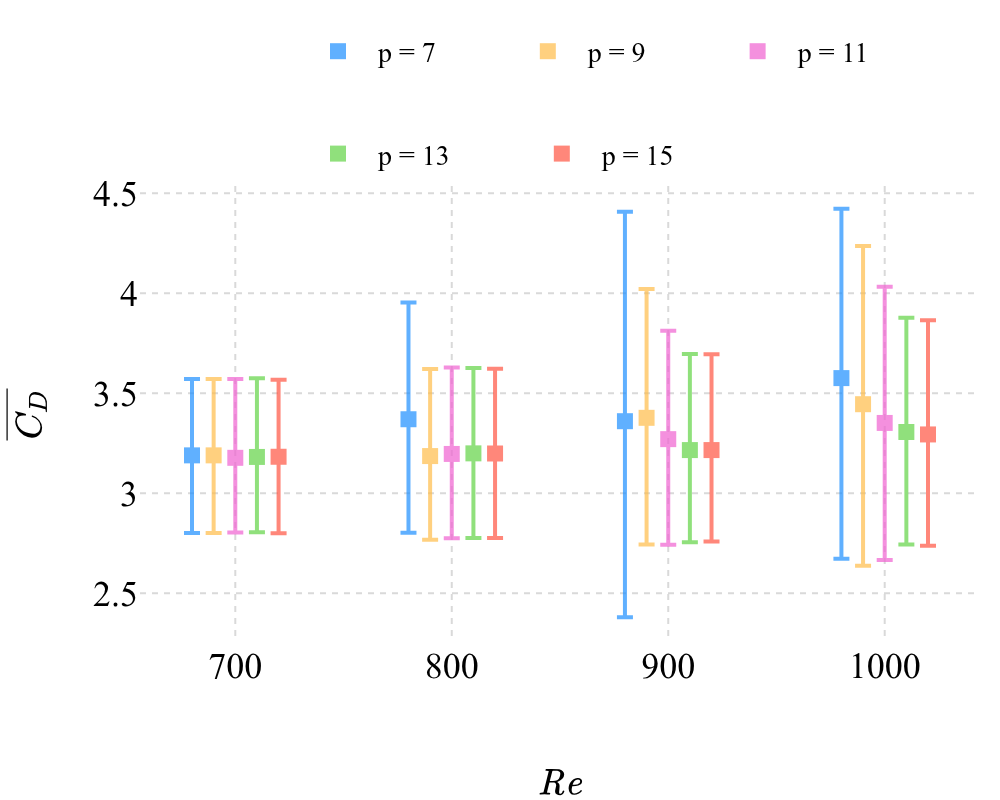}
    \end{subfigure}
    \begin{subfigure}[t]{0.49\textwidth}
        \centering
        \includegraphics[width=\textwidth]{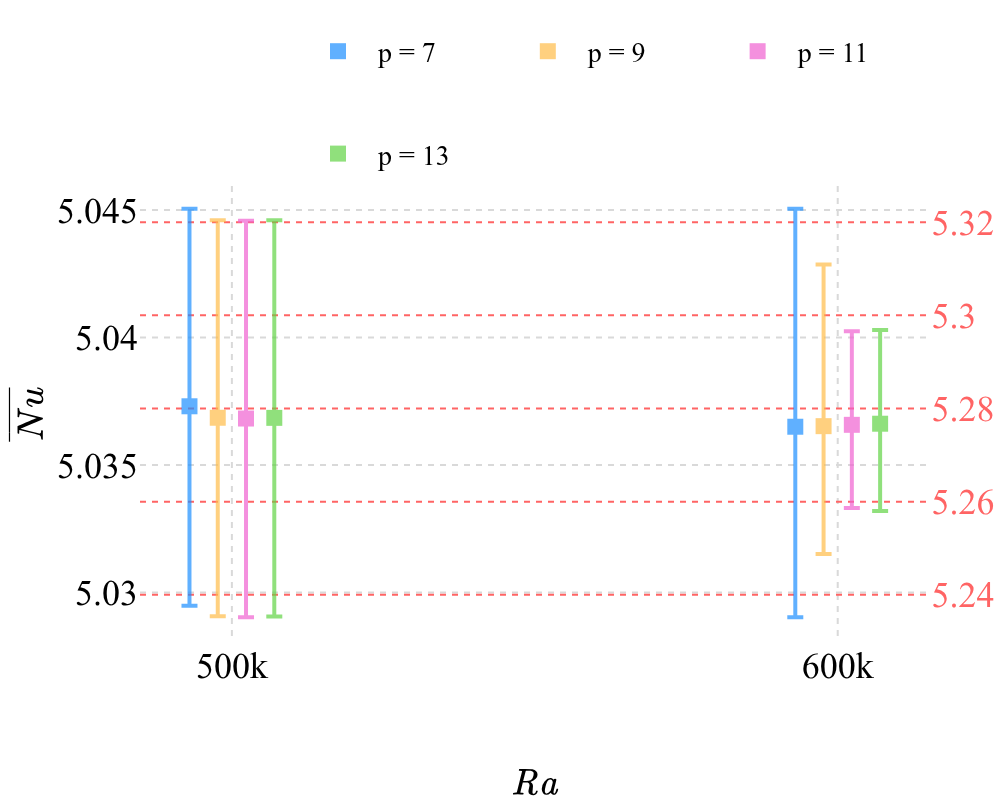}
    \end{subfigure}
    \caption{Effect of the polynomial order on the drag coefficient across different $\Rey$ (left) and on the Nusselt number across different $\Ray$ (right). Markers represent the mean values while the bars represent maximum and minimum values along the limit cycle. For $\Ray=600k$, results are shown on the right y-axis (red).}
    \label{fig:p_refinement}
\end{figure}

In order to ensure that simulations are properly converged, a p-refinement study is conducted for all systems investigated in this work as shown in Fig.~\ref{fig:p_refinement}. For the cylinder case, simulations with $\Rey \in [700, 1000]$ are performed and the drag coefficient is computed after initial transients. Results show that for $\Rey \leq 800$ a 9th order polynomial is appropriate. However, a 13th order polynomial is necessary to ensure convergence for $\Rey > 800$. The same procedure is conducted for the cavity flow, where the mean Nusselt number is used to assess convergence. Results show that a 9th order polynomial is appropriate for $Ra<5\times 10^5$. For $\Ray\geq 5\times 10^5$, a 13th order polynomial is employed.


%

\section{Unstructured Meshes}
\label{app:unstructured_meshes}
\begin{figure}[pos=ht!]
    \centering
    \includegraphics[width=1\textwidth]{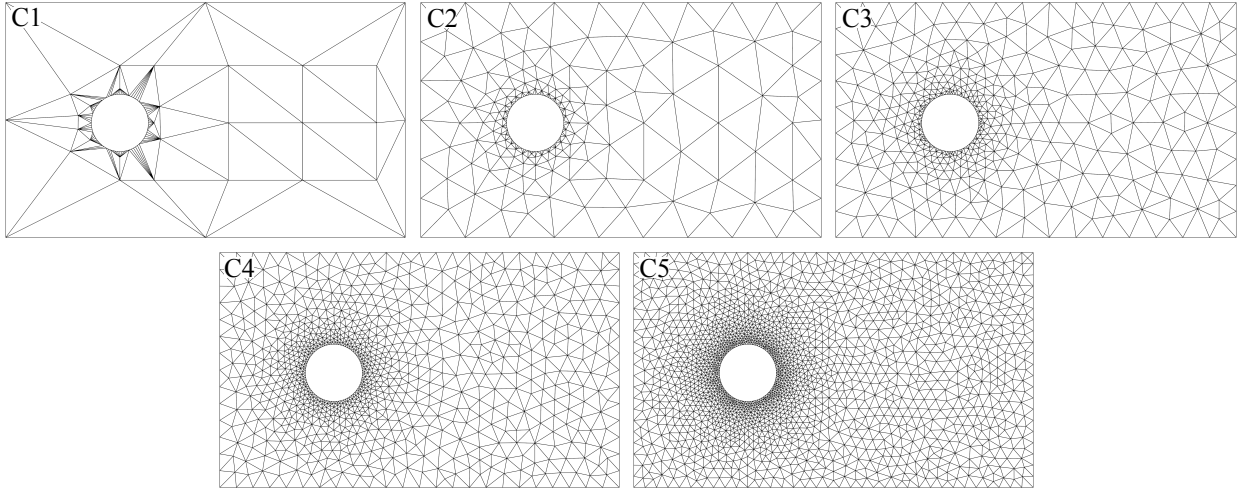}
    \caption{Meshes employed to trained models for the cylinder flow case.}
    \label{fig:cyl_meshes}
\end{figure}

\begin{figure}[pos=ht!]
    \centering
    \includegraphics[width=0.62\textwidth]{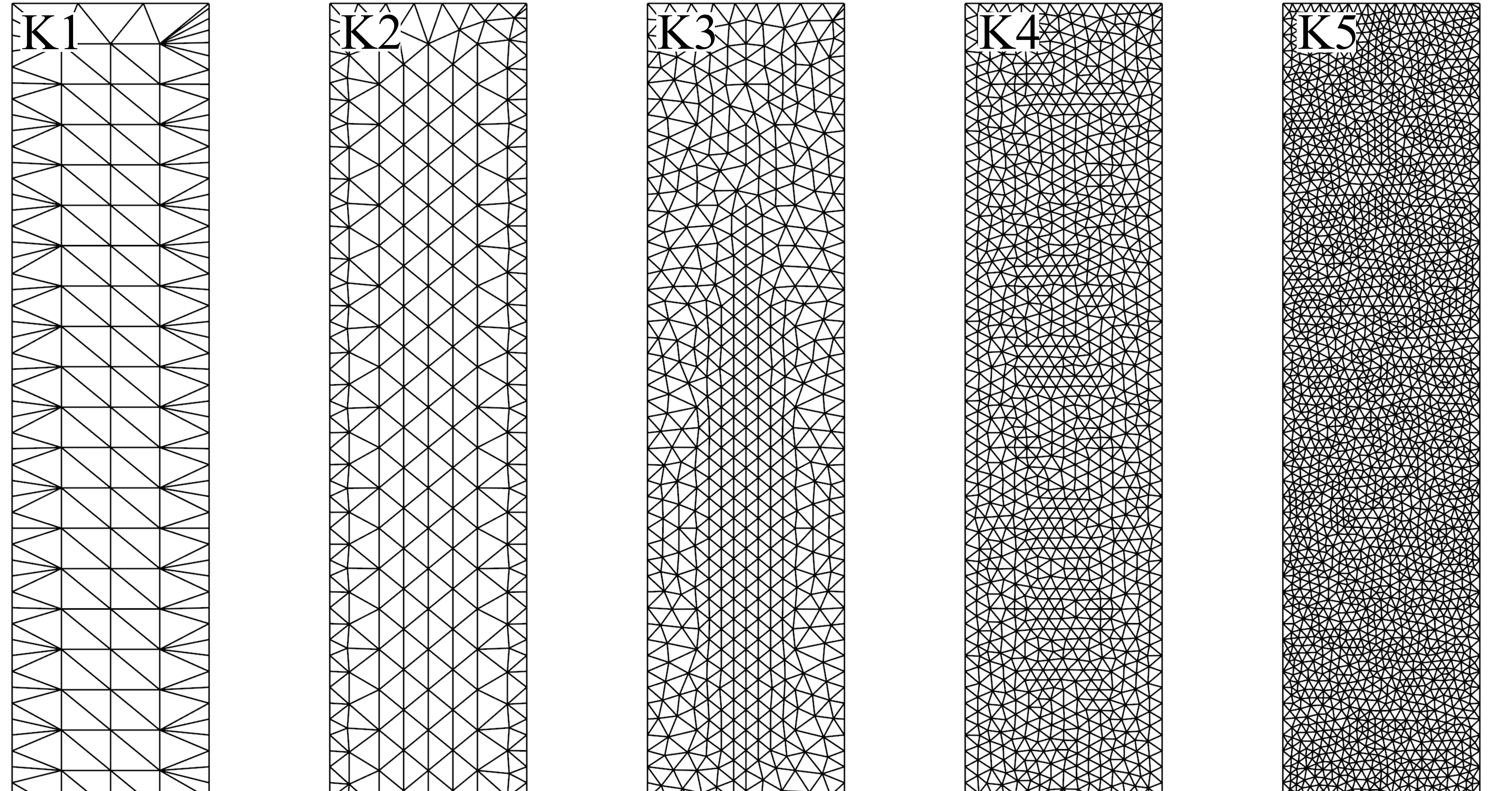}
    \caption{Meshes employed to trained models for the cavity flow case. Only half of the domain is being shown.}
    \label{fig:cav_meshes}
\end{figure}

\begin{figure}[pos=ht!]
    \centering
    \includegraphics[width=1\textwidth]{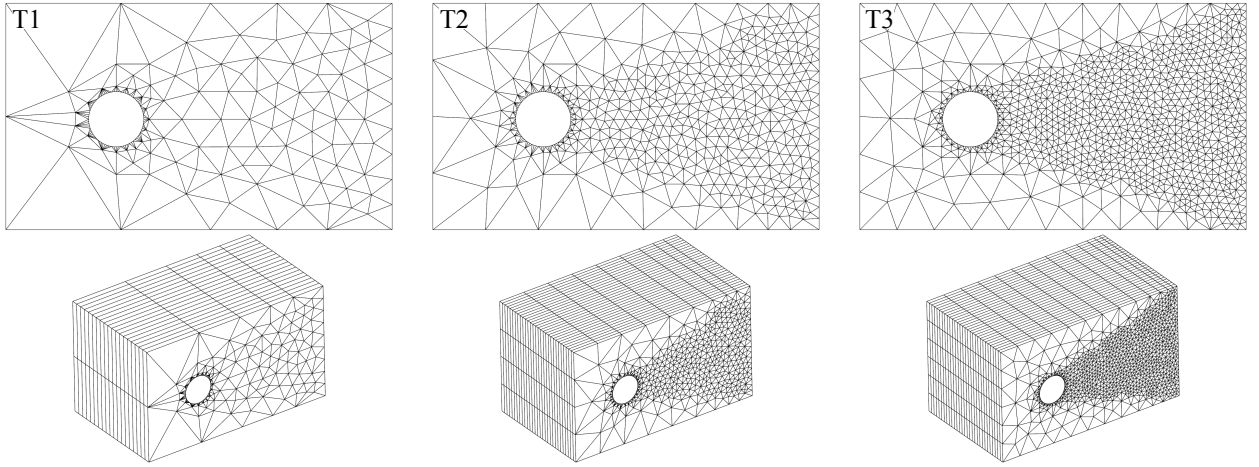}
    \caption{Meshes employed to trained models for the 3D cylinder flow case.}
    \label{fig:cyl3D_meshes}
\end{figure}

All meshes employed by the GNNs are created using Delaunay triangulation and are shown in Figures~\ref{fig:cyl_meshes} \ref{fig:cav_meshes} and \ref{fig:cyl3D_meshes}. For the 3D cylinder case, the mesh is extruded with 20, 25 and 30 slices in the spanwise direction for meshes T1, T2 and T3, respectively.

\section{Baseline Details}
\label{app:baselines}

The parametric DMD baseline uses the Reduced Koopman Operator Interpolation method \cite{li2021fourier}. It truncates the first 10 and 20 modes for the two-dimensional cylinder and cavity cases, respectively, obtained from the singular value decomposition (SVD) of the snapshot matrix. When considering the 3D cylinder case, the first 20 modes are used. For unseen parameter values, both the SVD modes and the reduced Koopman operator are linearly interpolated. The MLP baseline consists of 4 dense layers with ReLU activations in all layers except the output layer. The hidden layers use a latent dimension of 128, and training uses a batch size of 32.

The U-Net used in all experiments consists of an encoder that projects the inputs into a latent representation (64 and 32 channels for the 2D and 3D cases, respectively), followed by two downsampling blocks, two upsampling blocks, and a decoder. Skip connections concatenate the downsampled and upsampled latent features throughout the model. The encoder, downsampling, and upsampling blocks each consist of two stacked convolutional blocks. Each convolutional block comprises a convolutional layer with kernel size 3 and zero padding 1, followed by layer normalization and a ReLU activation. Each downsampling block applies max pooling with kernel size 2 before its convolutional block, whereas each upsampling block uses a 2D or 3D transposed convolution before its convolutional block. The decoder consists of a single 1x1 convolution. The number of channels is doubled after each downsampling block. For the 3D case, periodic convolutions are used in the spanwise direction.

The FNO is implemented using the publicly available Python packages from \cite{duruisseaux2025guide, kovachki2023neuraloperator}. The first 4 and 16 Fourier modes are retained in the horizontal and vertical directions, respectively. The FNO uses 4 layers with a latent dimension of 64. Both the U-Net and FNO models are trained with a batch size of 16. All deep learning models are trained to predict the difference between consecutive snapshots.

\printcredits

\bibliographystyle{cas-model2-names}

\bibliography{cas-refs}






\end{document}